\documentclass[review]{elsarticle}

\usepackage{lineno,hyperref}

\usepackage{booktabs}       
\usepackage{amsfonts}       
\usepackage{nicefrac}       
\usepackage{microtype}      
\usepackage{xcolor}         
\usepackage{amsmath}
\usepackage{amssymb}
\usepackage{amsthm}
\usepackage{subcaption}
\usepackage{bm}
\usepackage{threeparttable}
\usepackage{diagbox}
\usepackage{float}
\usepackage{multirow}
\newtheorem{theorem}{Theorem}

\newtheorem{remark}{Remark}

\usepackage{cases}
\usepackage{wrapfig}

\usepackage{url}
\usepackage[noend]{algpseudocode}
\usepackage{algorithmicx,algorithm}
\usepackage{graphicx}
\usepackage{tablefootnote}
\usepackage[font=scriptsize]{caption}
\allowdisplaybreaks[4]

\newtheorem{proposition}[theorem]{Proposition}

\journal{Neural Networks}

\biboptions{authoryear}

\begin{document}

\begin{frontmatter}

\title{SPIDE: A Purely Spike-based Method for Training Feedback Spiking Neural Networks}

\author[pku]{Mingqing Xiao}
\ead{mingqing\_xiao@pku.edu.cn}
\author[cuhk,sribd]{Qingyan Meng}
\ead{qingyanmeng@link.cuhk.edu.cn}
\author[ds]{Zongpeng Zhang}
\ead{zhangzongpeng@stu.pku.edu.cn}
\author[pku,pkuAI]{Yisen Wang}
\ead{yisen.wang@pku.edu.cn}
\author[pku,pkuAI,pengcheng]{Zhouchen Lin\corref{mycorrespondingauthor}}
\ead{zlin@pku.edu.cn}

\address[pku]{National Key Laboratory of General Artificial Intelligence, School of Intelligence Science and Technology, Peking University, China}
\address[cuhk]{The Chinese University of Hong Kong, Shenzhen, China}
\address[sribd]{Shenzhen Research Institute of Big Data, Shenzhen 518115, China}
\address[ds]{Center for Data Science, Academy for Advanced Interdisciplinary Studies, Peking University, China}
\address[pkuAI]{Institute for Artificial Intelligence, Peking University, China}
\address[pengcheng]{Peng Cheng Laboratory, China}
\cortext[mycorrespondingauthor]{Corresponding author}

\begin{abstract}

Spiking neural networks (SNNs) with event-based computation are promising brain-inspired models for energy-efficient applications on neuromorphic hardware. However, most supervised SNN training methods, such as conversion from artificial neural networks or direct training with surrogate gradients, require complex computation rather than spike-based operations of spiking neurons during training.
In this paper, we study spike-based implicit differentiation on the equilibrium state (SPIDE) that extends the recently proposed training method, implicit differentiation on the equilibrium state (IDE), for supervised learning with purely spike-based computation, which demonstrates the potential for energy-efficient training of SNNs.
Specifically, we introduce ternary spiking neuron couples and prove that implicit differentiation can be solved by spikes based on this design, so the whole training procedure, including both forward and backward passes, is made as event-driven spike computation, and weights are updated locally with two-stage average firing rates. Then we propose to modify the reset membrane potential to reduce the approximation error of spikes.
With these key components, we can train SNNs with flexible structures in a small number of time steps and with firing sparsity during training, and the theoretical estimation of energy costs demonstrates the potential for high efficiency. Meanwhile, experiments show that even with these constraints, our trained models can still achieve competitive results on MNIST, CIFAR-10, CIFAR-100, and CIFAR10-DVS. Our code is available at \url{https://github.com/pkuxmq/SPIDE-FSNN}.
\end{abstract}


\end{frontmatter}

\section{Introduction}
\label{introduction}

Spiking neural networks (SNNs) are brain-inspired models that transmit spikes between neurons for event-driven energy-efficient computation. SNNs can be implemented with less energy on neuromorphic hardware~\citep{akopyan2015truenorth,davies2018loihi,pei2019towards,roy2019towards}, which is regarded as the third generation of neural network models~\citep{maass1997networks} and is gaining increasing attention as an alternative to artificial neural networks (ANNs).

Different from ANNs, however, directly supervised training of SNNs is a hard problem due to the complex spiking neuron model which is discontinuous. To handle this problem, converting ANNs to SNNs~\citep{hunsberger2015spiking,rueckauer2017conversion,sengupta2019going,rathi2019enabling,deng2021optimal,yan2021near}, or many direct SNN training methods~\citep{wu2018spatio,bellec2018long,jin2018hybrid,shrestha2018slayer,wu2019direct,neftci2019surrogate,zhang2019spike,kim2020unifying,zheng2020going,bohte2002error,zhang2020temporal,kim2020unifying,xiao2021training,meng2022training} have been proposed to incorporate deep learning into SNNs~\citep{tavanaei2019deep}. While these methods can partly solve the problems of unsatisfactory performance or high latency, they require complex computation for gradient calculation or approximation rather than the same spike-based computation as the inference process. They aim at training SNNs with general computational operations and deploying trained models for energy-efficient inference with spiking operations.

As a different direction compared with these methods, it is an important problem to consider if the training process can also take advantage of the spiking nature of common spiking neurons so that the computation of inference and training is consistent and the training process can also share the energy efficiency with event-driven computation. The consistency of the inference and training computation may also pave a path for directly training SNNs on neuromorphic hardware instead of training models on commonly used computational units and deploying them on chips.
A few previous works try to train SNNs with spikes~\citep{guerguiev2017towards,neftci2017event,samadi2017deep,o2016deep,thiele2019spiking,thiele2019spikegrad}. They either are based on direct feedback alignment (DFA)~\citep{nokland2016direct} and perform poorly, or require special neuron models~\citep{thiele2019spiking,thiele2019spikegrad} that are hardly supported. Besides, they only focus on feedforward network structures imitated from ANNs, which ignores feedback connections that are ubiquitous in the human brain and enable neural networks to be shallower and more efficient~\citep{kubilius2019brain,xiao2021training}. Actually, feedback structures suit SNNs more since SNNs will naturally compute with multiple time steps, which can reuse representations and avoid additional uneconomical costs to unfold along the time as in ANNs~\citep{xiao2021training,kim2022neural}. So training algorithms for feedback SNNs, which may also be degraded for feedforward structures by taking feedback as zero, is worth more exploration.

An ideal SNN training method should tackle the common problems, be suitable for flexible structures (feedforward or feedback), and be with spike-based computation for efficiency and high neuromorphic plausibility.
The implicit differentiation on the equilibrium state (IDE) method~\citep{xiao2021training}, which is recently proposed to train feedback spiking neural networks (FSNNs), is a promising method that may be generalized to spike-based learning for the requirement. They derive that the forward computation of FSNNs converges to an equilibrium state, which follows a fixed-point equation. Based on it, they propose to train FSNNs by implicit differentiation on this equation, which tackles the common difficulties for SNN training including non-differentiability and large memory costs, and has interesting local update properties. In their method, however, they leverage general root-finding methods rather than spike-based computation to solve implicit differentiation.

In this work, we extend the IDE method to spike-based IDE (SPIDE), which fulfills our requirements and has great potential for energy-efficient training with spike-based computation. We introduce ternary spiking neuron couples and propose to solve implicit differentiation by spikes based on them. Our method is also applicable to feedforward structures by setting the feedback connection as zero. In practice, however, it may require long time steps to stabilize the training with spikes due to the approximation error for gradients. So we further dive into the approximation error from the statistical perspective and propose to simply adjust the reset potential of SNNs to achieve an unbiased estimation of gradients and reduce the estimation variance of SNN computation. With these methods, we can train our models in a small number of time steps, which can further improve the energy efficiency as well as the latency. This demonstrates the strong power of spike-based computation for training SNNs. Our contributions include:
\begin{enumerate}
    \item We propose the SPIDE method which is the first to train high-performance SNNs by spike-based computation with common neuron models. Specifically, we propose ternary spiking neuron couples and prove that implicit differentiation for gradient calculation can be solved by spikes based on this design. Our method is applicable to both feedback and feedforward structures.
    \item We theoretically analyze the approximation error of solving implicit differentiation by spikes, and propose to modify the reset potential to remove the approximation bias and reduce the estimation variance, which enables training in a small number of time steps.
    \item Experiments show the low latency and firing sparsity during training, and the theoretical estimation of energy costs demonstrates the great potential for energy-efficient training of SNNs with spike-based computation. The performance on MNIST, CIFAR-10, CIFAR-100 and CIFAR10-DVS are competitive as well.
\end{enumerate}

\section{Related Work}

\paragraph{SNN Training Methods}

\begin{table*} [ht!]\scriptsize
	\tabcolsep=0.5mm
	\caption{Comparison of different supervised SNN training methods with respect to performance, latency, structure flexibility, neuron model, spike-based or not, and neuromorphic plausibility.}
	\hspace{-6em}
	\begin{threeparttable}
	\begin{tabular}{ccccccc}
		\toprule[1pt]
		Method & High Perform. & Low Latency & Struc. Flexi. & Common Neuron Model & Spike-based & Neuro. Plaus. \\
		\midrule[0.5pt]
		ANN-to-SNN & \checkmark & \checkmark\tnote{*} & $\times$ & \checkmark & $\times$ & Low\\
		BPTT with Surrogate Gradients & \checkmark & \checkmark & \checkmark & \checkmark & $\times$ & Low\\
		DFA with Spikes & $\times$ & ? & $\times$ & \checkmark & \checkmark & High\\
		SpikeGrad~\citep{thiele2019spikegrad} & \checkmark & ? & $\times$ & $\times$ & \checkmark & Medium\\
		IDE~\citep{xiao2021training} & \checkmark & \checkmark & \checkmark & \checkmark & $\times$ & Low\\
		\midrule[0.5pt]
		\textbf{SPIDE (ours)} & \checkmark & \checkmark & \checkmark & \checkmark & \checkmark & High\\
		\bottomrule[1pt]
	\end{tabular}
	\begin{tablenotes}
       \scriptsize
       \item[*] Typical ANN-to-SNN methods require high latency. Some recent state-of-the-art works have largely improved this.
     \end{tablenotes}
	\end{threeparttable}
	\label{method comparison}
\end{table*}

Early works seek biologically inspired methods to train SNNs, e.g. spike-time dependent plasticity (STDP)~\citep{diehl2015unsupervised,kheradpisheh2018stdp} or reward-modulated STDP~\citep{legenstein2008learning}.
Since the rise of successful ANNs, several works try to convert trained ANNs to SNNs to obtain high performance~\citep{hunsberger2015spiking,rueckauer2017conversion,sengupta2019going,rathi2019enabling,deng2021optimal,yan2021near,li2021free,stockl2021optimized}. However, they typically suffer from extremely large time steps (some recent state-of-the-art works largely improved this~\citep{li2021free,bu2021optimal,meng2022trainingnn}) and their structures are limited in the scope of ANNs. Others try to directly train SNNs by backpropagation through time (BPTT) and use surrogate derivative for discontinuous spiking functions~\citep{lee2016training,wu2018spatio,bellec2018long,jin2018hybrid,shrestha2018slayer,wu2019direct,zhang2019spike,neftci2019surrogate,zheng2020going,Fang_2021_ICCV,li2021differentiable,fang2021deep,deng2021temporal} or compute gradient with respect to spiking times~\citep{bohte2002error,zhang2020temporal,kim2020unifying}. However, they suffer from approximation errors and large training memory costs, and their optimization with surrogate gradients is not well guaranteed theoretically. \citet{xiao2021training} propose the IDE method to train feedback spiking neural networks, which decouples the forward and backward procedures and avoids the common SNN training problems. Tandem learning~\citep{wu2021tandem}, ASF~\citep{wu2021training}, and DSR~\citep{meng2022training} also share a similar thought of calculating gradients through spike rates, with the main focus only on feedforward networks with closed-form transformations between successive layers instead of implicit fixed-point equations at equilibrium states. However, all these methods require complex computation during training rather than spike-based operations. A few works focusing on training SNNs with spikes either are based on feedback alignment and limited in simple datasets~\citep{guerguiev2017towards,neftci2017event,samadi2017deep,o2016deep}, or require special neuron models that require consideration of accumulated spikes for spike generation~\citep{thiele2019spiking,thiele2019spikegrad}, which is hardly supported in practice. And they are only applicable to feedforward architectures. Instead, we are the first to leverage spikes with common neuron models to train high-performance SNNs with feedback or feedforward structures. The comparison of different methods is illustrated in Table~\ref{method comparison}.

\paragraph{Equilibrium of Neural Networks}
The equilibrium of neural networks is first considered by energy-based models, e.g. Hopfield Network~\citep{hopfield1982neural,hopfield1984neurons}, which view the dynamics of feedback neural networks as minimizing an energy function towards an equilibrium of the local minimum of the energy. With the energy function, recurrent backpropagation~\citep{almeida1987learning,pineda1987generalization} and more recent equilibrium propagation (EP)~\citep{scellier2017equilibrium} are proposed to train neural networks. These methods rely on the definition of the energy function and are hardly competitive with deep neural networks.
With the steady state of equilibrium, the contrastive Hebbian learning method~\citep{xie2003equivalence} is also proposed to train neural networks, which is composed of a two-phase anti-Hebbian and Hebbian procedure and is shown equivalent to backpropagation in the limit of weak feedback. \citet{detorakis2019contrastive} propose a variant of this method with random feedback weights which is similar to feedback alignment. However, these methods hardly reach the high performance of backpropagation.
Recently, a new branch of deep equilibrium models~\citep{bai2019deep,bai2020multiscale} is proposed. They interpret the computation of weight-tied deep ANNs as solving a fixed-point equilibrium point, and correspondingly propose implicit models which are defined by solving the equilibrium equations and trained by solving implicit differentiation. Their equilibrium is defined by fixed-point equations rather than energy functions and they can achieve superior performance. These works focus on ANNs rather than SNNs, except that \citet{mesnard2016towards}, \citet{o2019training} and \citet{martin2021eqspike} generalize the idea of EP to SNNs. They follow the energy-based EP method to approximate the gradients, which is different from the equilibrium and the training method in this work, and they are limited to tiny network scales and simple datasets. As for the equilibrium of SNNs, \citet{zhang2018plasticity,zhang2018brain} consider the equilibrium state of the membrane potential as an unsupervised learning part, \citet{li2020minimax} and \citet{mancoo2020understanding} consider the equilibrium from the perspective of solving constrained optimization problems, and \citet{xiao2021training} draw inspiration from deep equilibrium models to view feedback SNNs as evolving along time to an equilibrium state following a fixed-point equation and propose to train FSNNs by implicit differentiation on the equilibrium state (IDE). These methods do not consider training SNNs with spike-based computation. Differently, this work extends IDE to spike-based IDE, which extends the thought of equilibrium of spikes to solving implicit differentiation and enables the whole training procedure to be based on spike computation with common neuron models, providing the potential for energy-efficient training of SNNs.

\section{Preliminaries}

We first introduce preliminaries about spiking neural network models and the IDE training method~\citep{xiao2021training}.

\subsection{Spiking Neural Network Models}

Spiking neurons draw inspiration from the human brain to communicate with each other by spikes. Each neuron integrates information from input spike trains by maintaining a membrane potential through a differential equation and generates an output spike once the membrane potential exceeds a threshold, following which the membrane potential is reset to the reset potential. We consider the commonly used integrate and fire (IF) and leaky integrate and fire (LIF) neuron models, whose general discretized computational form is:
\begin{equation}
    \left\{
    \begin{aligned}
        &u_i\left[t + 1\right] = \lambda(u_i[t] - (V_{th}-u_{reset})s_i[t]) + \sum_j w_{ij}s_j[t] + b_i,\\
        &s_i[t + 1] = H(u_i\left[t+1\right] - V_{th}),\\
    \end{aligned}
    \right.
    \label{eq.snn}
\end{equation}
where $u_i[t]$ is the membrane potential of neuron $i$ at time step $t$, $s_i[t]$ is the binary output spike train of neuron $i$, $w_{ij}$ is the connection weight from neuron $j$ to neuron $i$, $b_i$ is bias, $H$ is the Heaviside step function, $V_{th}$ is the firing threshold, $u_{reset}$ is the reset potential, and $\lambda=1$ for the IF model while $\lambda < 1$ is a leaky term for the LIF model. We use subtraction as the reset operation. $u_{reset}$ is usually taken as $0$ in previous work, while we will reconsider it in Section~\ref{subsec:error}. We mainly consider the IF neuron model by default and will demonstrate that our method is applicable to the LIF model as well.

\subsection{Background about the IDE Training Method}

Due to the complex spiking neuron model which is discontinuous, directly supervised training of SNNs is a hard problem, since the explicit computation is non-differentiable and therefore backpropagation along the forward computational graph is problematic. The IDE training method~\citep{xiao2021training} considers another approach to calculate gradients that does not rely on the exact reverse of the forward computation, which avoids the problem of non-differentiability as well as large memory costs by BPTT methods with surrogate gradients. Specifically, the IDE training method first derives that the (weighted) average firing rate of FSNN computation with common neuron models would gradually evolve to an equilibrium state along time, which follows a fixed-point equation. Then by viewing the forward computation of FSNN as a black-box solver for this equation, and applying implicit differentiation on the equation, gradients can be calculated only based on this equation and the (weighted) average firing rate during the forward computation rather than the exact forward procedure. Therefore, the forward and backward procedures are decoupled and the non-differentiability is avoided.

We briefly introduce the conclusion of equilibrium states in Section~\ref{equilibrium states} and the IDE method in Section~\ref{ide training method}.

\subsubsection{Equilibrium States of FSNNs}\label{equilibrium states}

\citet{xiao2021training} derive that the (weighted) average rate of spikes during FSNN computation with common neuron models would converge to an equilibrium state following a fixed-point equation given convergent inputs.
We first focus on the conclusions with the discrete IF model under both single-layer and multi-layer feedback structures. The single-layer structure has one hidden layer of neurons with feedback connections on this layer. The update equation of membrane potentials after reset is (let $V_u=V_{th}-u_{reset}$):
\begin{equation}
    \mathbf{u}[t+1] = \mathbf{u}[t] + \mathbf{W}\mathbf{s}[t] + \mathbf{F}\mathbf{x}[t+1] + \mathbf{b} - V_u\mathbf{s}[t+1],
    \label{eq.singlelayer}
\end{equation}
where $\mathbf{u}[t]$ and $\mathbf{s}[t]$ are the vectors of membrane potentials and spikes of these neurons, $\mathbf{x}[t]$ is the input at time step $t$, $\mathbf{W}$ is the feedback weight matrix, and $\mathbf{F}$ is the weight matrix from inputs to these neurons. The average input and average firing rate are defined as $\mathbf{\overline{x}}[t]=\frac{1}{t}\sum_{\tau=1}^t \mathbf{x}[\tau]$ and $\bm{\alpha}[t]=\frac{1}{t}\sum_{\tau=1}^t \mathbf{s}[\tau]$, respectively. Define $\sigma(x)=\min(1, \max(0, x))$.

The equilibrium state of the single-layer FSNN is described as~\citep{xiao2021training}: If the average inputs converge to an equilibrium point $\mathbf{\overline{x}}[t]\rightarrow \mathbf{x}^*$, and there exists $\gamma<1$ such that $\lVert \mathbf{W} \rVert_2 \leq \gamma V_{th}$, then the average firing rates of FSNN with discrete IF model will converge to an equilibrium point $\bm{\alpha}[t]\rightarrow \bm{\alpha}^*$, which satisfies the fixed-point equation $\bm{\alpha}^* = \sigma\left(\frac{1}{V_{th}}\left(\mathbf{W}\bm{\alpha}^*+\mathbf{F}\mathbf{x}^*+\mathbf{b}\right)\right)$.
Note that they take $u_{reset}=0$ in this conclusion, if we consider nonzero $u_{reset}$, the constraint and the fixed-point equation should be $\lVert \mathbf{W} \rVert_2 \leq \gamma (V_{th}-u_{reset})$ and $\bm{\alpha}^* = \sigma\left(\frac{1}{V_{th}-u_{reset}}\left(\mathbf{W}\bm{\alpha}^*+\mathbf{F}\mathbf{x}^*+\mathbf{b}\right)\right)$.

The multi-layer structure incorporates more non-linearity into the equilibrium fixed-point equation, which has multiple layers with feedback connections from the last layer to the first layer. The update equations of membrane potentials after reset are expressed as:
\begin{equation}
\small
    \left\{
    \begin{aligned}
        \mathbf{u}^1[t + 1] = \mathbf{u}^1[t] &+ \mathbf{W}^1\mathbf{s}^N[t] + \mathbf{F}^1\mathbf{x}[t+1]+\mathbf{b}^1-V_u\mathbf{s}^1[t+1],\\
        \mathbf{u}^{l}[t + 1] = \mathbf{u}^{l}[t] &+ \mathbf{F}^{l}\mathbf{s}^{l-1}[t+1]+\mathbf{b}^{l}-V_u\mathbf{s}^{l}[t+1], ( l=2,\cdots, N ).\\
    \end{aligned}
    \right.
    \label{eq.multilayer}
\end{equation}

The equilibrium state of the multi-layer FSNN with $u_{reset}$ is described as~\citep{xiao2021training}: If the average inputs converge to an equilibrium point $\mathbf{\overline{x}}[t]\rightarrow \mathbf{x}^*$, and there exists $\gamma<1$ such that $\lVert \mathbf{W}^1\rVert_2\lVert \mathbf{F}^N\rVert_2\cdots\lVert \mathbf{F}^2\rVert_2 \leq \\ \gamma {(V_{th}-u_{reset})}^N$, then the average firing rates of multi-layer FSNN with discrete IF model will converge to equilibrium points $\bm{\alpha}^l[t]\rightarrow {\bm{\alpha}^l}^*$, which satisfy the fixed-point equations ${\bm{\alpha}^1}^* = f_1\left(f_N\circ\cdots\circ f_2({\bm{\alpha}^1}^*), \mathbf{x}^*\right)$ and ${\bm{\alpha}^{l+1}}^*=f_{l+1}({\bm{\alpha}^l}^*)$, where $f_1(\bm{\alpha}, \mathbf{x})=\sigma\left(\frac{1}{V_{th}-u_{reset}}(\mathbf{W}^1\bm{\alpha}+\mathbf{F}^1\mathbf{x}+\mathbf{b}^1)\right)$ and $f_{l}(\bm{\alpha}) = \sigma\left(\frac{1}{V_{th}-u_{reset}}(\mathbf{F}^{l}\bm{\alpha}+\mathbf{b}^{l})\right)$.

As for the LIF model, the weighted average firing rate will be considered to represent information: the weighted average input and weighted average firing rate are defined as $\mathbf{\hat{x}}[t]=\frac{\sum_{\tau=1}^t \lambda^{t-\tau}\mathbf{x}[\tau]}{\sum_{\tau=1}^t \lambda^{t-\tau}}$ and $\hat{\bm{\alpha}}[t]=\frac{\sum_{\tau=1}^t \lambda^{t-\tau}\mathbf{s}[\tau]}{\sum_{\tau=1}^t \lambda^{t-\tau}}$, respectively. \citet{xiao2021training} has derived the equilibrium states of FSNN under the LIF model as well. The conclusions are similar to the IF model and the equilibrium states follow the same fixed-point equation, except that weighted average firing rates (inputs) are considered and there would be bounded random error for the equilibrium state. We can view it as an approximate solver for the equilibrium state.

\subsubsection{IDE Training Method}\label{ide training method}

Based on the equilibrium states in Section~\ref{equilibrium states}, we can train FSNNs by calculating gradients with implicit differentiation~\citep{xiao2021training}. Let $\bm{\alpha}=f_{\theta}(\bm{\alpha})$ denote the fixed-point equation of the equilibrium state which is parameterized by $\theta$, $g_{\theta}(\bm{\alpha})=f_{\theta}(\bm{\alpha})-\bm{\alpha}$, and let $\mathcal{L}(\bm{\alpha}^*)$ denote the objective function with respect to the equilibrium state $\bm{\alpha}^*$. The implicit differentiation satisfies $\left(I-\frac{\partial f_{\theta}(\bm{\alpha}^*)}{\partial \bm{\alpha}^*}\right)\frac{\mathrm{d}\bm{\alpha}^*}{\mathrm{d} \theta}=\frac{\partial f_{\theta}(\bm{\alpha}^*)}{\partial \theta}$~\citep{bai2019deep} (we follow the numerator layout convention for derivatives).
Therefore, the differentiation of $\mathcal{L}(\bm{\alpha}^*)$ for parameters can be calculated based on implicit differentiation as:
\begin{equation}
    \frac{\partial \mathcal{L}(\bm{\alpha}^*)}{\partial \theta} = -\frac{\partial \mathcal{L}(\bm{\alpha}^*)}{\partial \bm{\alpha}^*} \left(J_{g_{\theta}}^{-1}\vert_{\bm{\alpha}^*}\right) \frac{\partial f_{\theta}(\bm{\alpha}^*)}{\partial \theta},
    \label{eq:gradients implicit differentiation}
\end{equation}
where $J_{g_{\theta}}^{-1}\vert_{\bm{\alpha}^*}$ is the inverse Jacobian of $g_{\theta}$ evaluated at $\bm{\alpha}^*$. The calculation of inverse Jacobian can be avoided by solving an alternative linear system~\citep{bai2019deep,bai2020multiscale,xiao2021training}:
\begin{equation}
    \left(J_{g_{\theta}}^\top\vert_{\bm{\alpha}^*}\right)\bm{\beta}+\left(\frac{\partial \mathcal{L}(\bm{\alpha}^*)}{\partial \bm{\alpha}^*}\right)^\top=0.
    \label{eq:linear equation}
\end{equation}
Note that a readout layer after the last layer of neurons will be constructed for output~\citep{xiao2021training}, which is composed of neurons that will not spike and will do classification based on accumulated membrane potentials (e.g. realized by a very large threshold). Then the output $o$ is equivalent to a linear transformation on the approximate equilibrium state, i.e. $\mathbf{o}=\mathbf{W}^o \bm{\alpha}^*+\mathbf{b}^o$, and the loss will be calculated between $\mathbf{o}$ and labels $\mathbf{y}$ with a common criterion such as cross-entropy. Then the gradient on the equilibrium state can be calculated. For the solution of implicit differentiation, \citet{xiao2021training} follow \citet{bai2019deep,bai2020multiscale} to leverage root-finding methods, while we will solve it by spike-based computation, as will be derived in Section~\ref{sec:method}. We consider that the (weighted) average firing rates $\bm{\alpha}[T]$ during the forward computation of FSNNs at time step $T$ roughly reach the equilibrium state. Then by substituting $\bm{\alpha}^*$ by $\bm{\alpha}[T]$ in the above equations, gradients for the parameters can be calculated only with $\bm{\alpha}[T]$ and the equation, and we calculate them based on spikes. With the gradients, first-order optimization methods such as SGD~\citep{rumelhart1986learning} and its variants can be applied to update parameters.

\section{Spike-based Implicit Differentiation on the Equilibrium State}\label{sec:method}

\begin{figure}[h!]
    \centering
    \begin{subfigure}[]{\textwidth}
    \includegraphics[width=\textwidth]{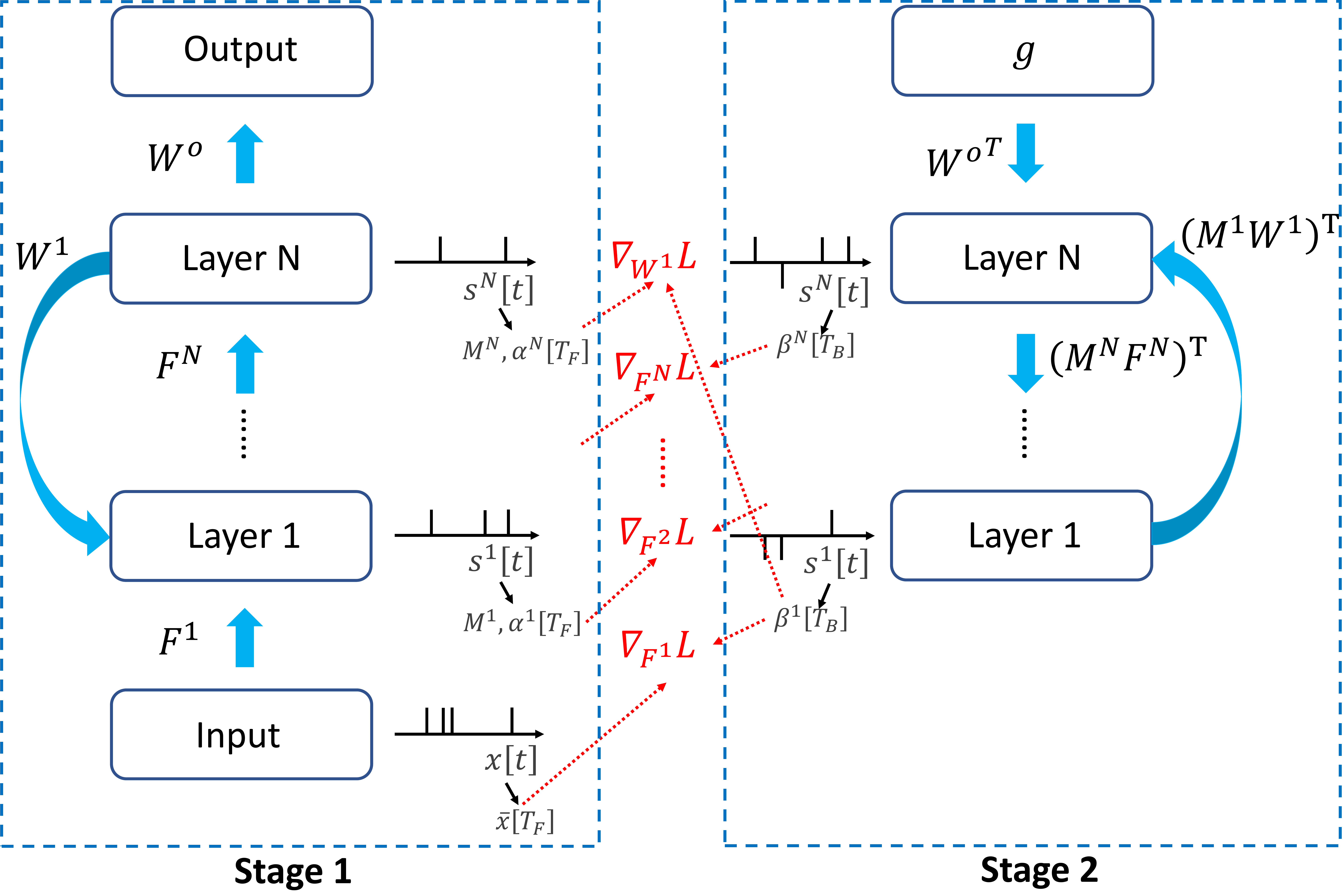}
    \label{spide}
    \vspace{-6mm}
    \caption{SPIDE}
    \end{subfigure}
    \begin{subfigure}[]{0.427\textwidth}
    \includegraphics[width=\textwidth]{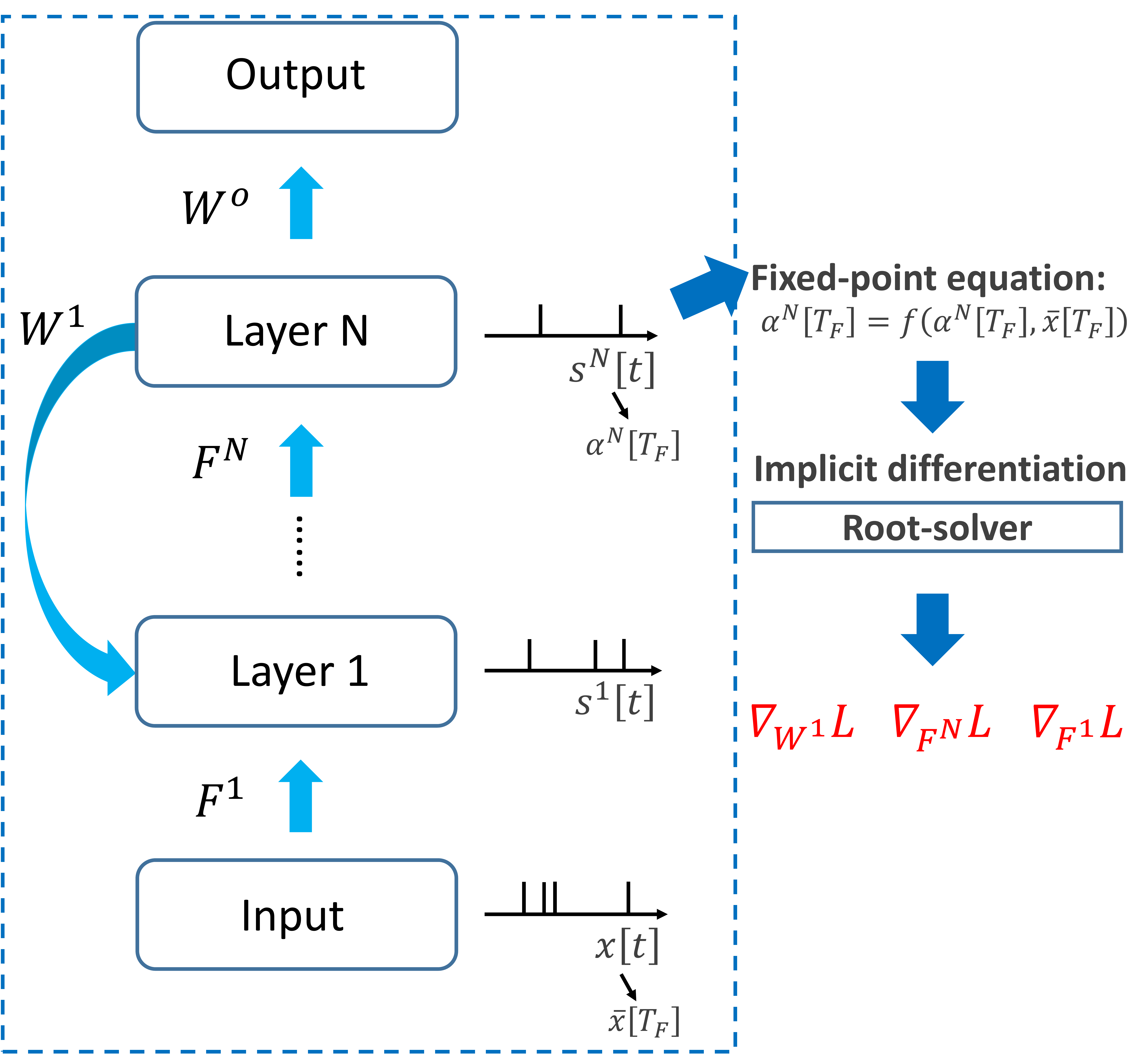}
    \label{ide}
    \vspace{-6mm}
    \caption{IDE}
    \end{subfigure}
    \hspace{6mm}
    \begin{subfigure}[]{0.5\textwidth}
    \includegraphics[width=\textwidth]{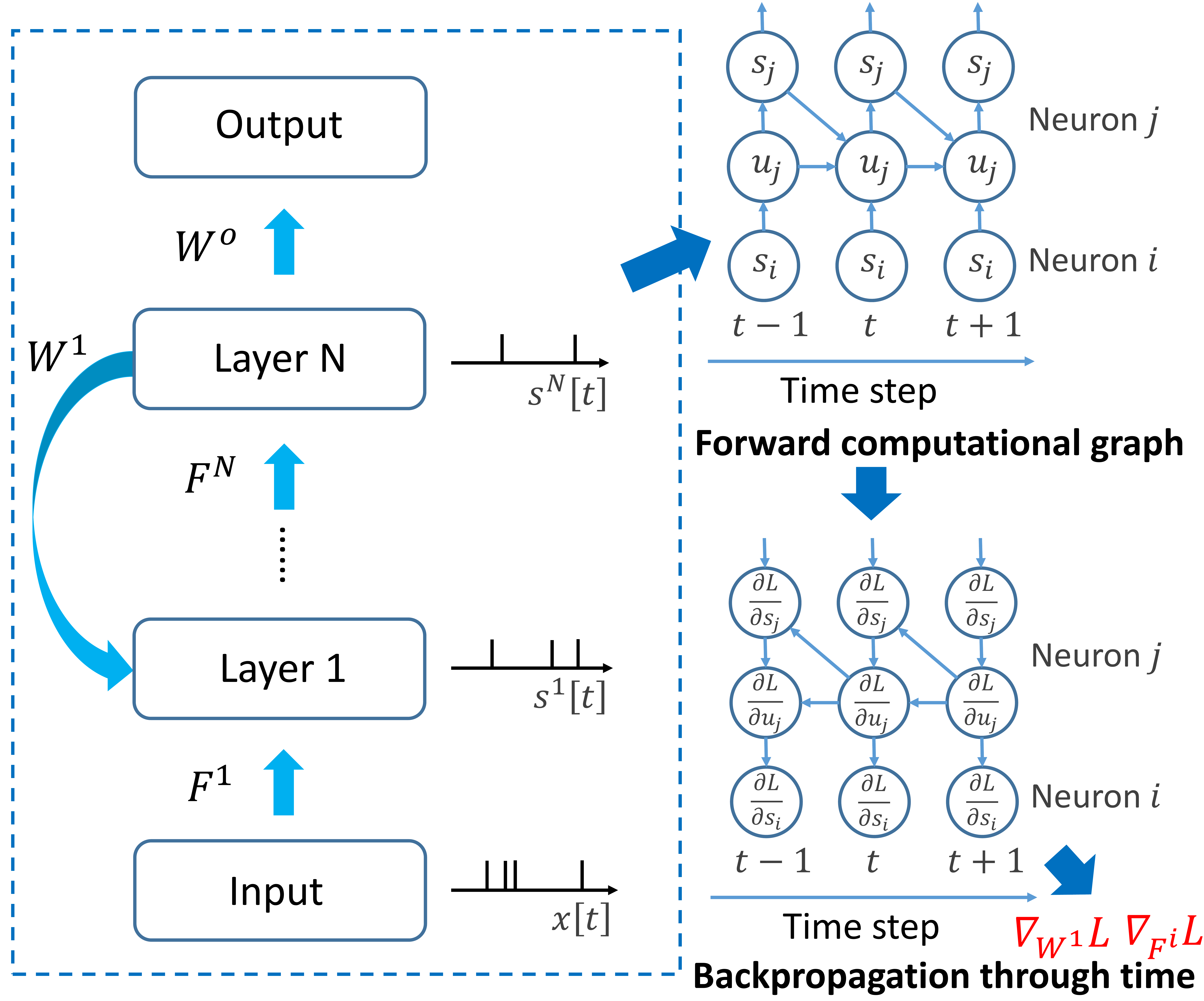}
    \label{bptt}
    \vspace{-7mm}
    \caption{BPTT with SG}
    \end{subfigure}
    \caption{A brief illustration of the overall process of the proposed method and comparison with previous methods. (a) The proposed SPIDE method. Both forward and backward stages are implemented by spike-based computation. (b) The IDE method. The backward stage requires root-solver and complex computation to solve for gradients. (c) The BPTT with surrogate gradient (SG) method. The backward stage requires backpropagation through the unfolded computational graph with complex computation rather than spike operations.}
    \label{fig: illustration}
\end{figure}

In this section, we present our SPIDE method that calculates the whole training procedure based on spikes. We first introduce ternary spiking neuron couples in Section~\ref{subsec: coupled neurons} and how to solve implicit differentiation in Section~\ref{subsec:solve id}. Then we theoretically analyze the approximation error and propose the improvement in Section~\ref{subsec:error}. Finally, a summary of the training pipeline is presented in Section~\ref{subsec:pipeline}. We also provide a brief illustration figure of the overall process in Figure~\ref{fig: illustration}.

\subsection{Ternary Spiking Neuron Couples}\label{subsec: coupled neurons}

The common spiking neuron model only generates spikes when the membrane potential exceeds a positive threshold, which limits the firing rate from representing negative information. To enable approximation of possible negative values for implicit differentiation calculation in Section~\ref{subsec:solve id}, we require negative spikes, whose expression could be:
\begin{equation}
\small
    s_i[t+1]=T\left(u_i[t+1], V_{th}\right)=\left\{\begin{aligned}
    &1, &&u_i[t+1]> V_{th}\\
    &0, &&\lvert u_i[t+1]\rvert\leq V_{th}\\
    &-1, &&u_i[t+1]<-V_{th}\\
\end{aligned}\right.,
\label{eq.ternary}
\end{equation}
and the reset is the same as usual: $u_i[t+1]-(V_{th}-u_{reset})s_i[t+1]$.
Direct realization of such ternary output, however, may be not supported by common neuromorphic computation of SNNs.

\begin{wrapfigure}{rht}{0.45\textwidth}
\vspace{-14mm}
    \begin{center}
        \includegraphics[width=0.44\textwidth]{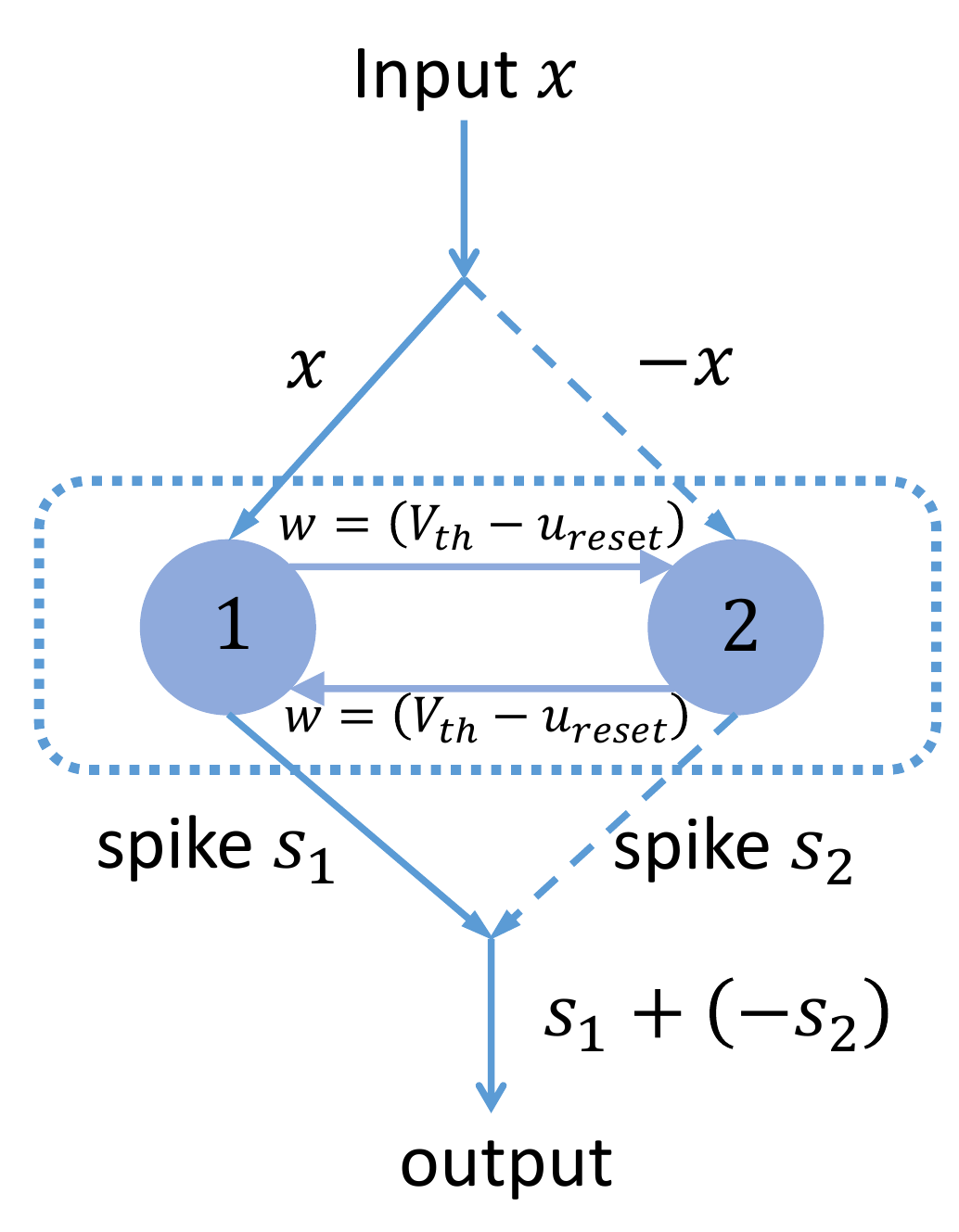}
    \end{center}
    \caption{Illustration of ternary spiking neuron couples.}
    \label{fig: coupled neurons}
\end{wrapfigure}
We propose to leverage two coupled common neurons to realize this computation. As illustrated in Figure~\ref{fig: coupled neurons}, the two coupled neurons with the common IF model (Eq.~(\ref{eq.snn})) receive opposite inputs and output opposite spikes, which aim to deal with positive information and negative information with spikes, respectively.
They should share a reset operation in order to accord with Eq.~(\ref{eq.ternary}), which can be realized by the connections between them: as we use subtraction as the reset operation, the connection whose weight equals $V_{th}-u_{reset}$ enables one neuron to reset the other equivalently. To see how this works, consider the condition that the accumulated membrane potential of neuron 1 reaches $V_{th}$, then neuron 1 would generate a spike and reset, and the output is this positive spike. At the same time, the membrane potential of neuron 2 is $-V_{th}$ and the neuron will not fire and reset, but the spike from neuron 1 will reset it to $-u_{reset}$, which accords with our desired reset for ternary output. Similarly, if the inputs are negative, neuron 2 will generate a spike which will be treated negative as output, and both neurons are reset. For the operation of taking negative, one solution is to enable the reverse operation on hardware; another is to reconnect neuron 2 with other neurons while taking the weight negative to that of neuron 1. As for most existing neuromorphic hardware, the double connections to the two coupled neurons are more feasible, and this implies that twice the synaptic operations are needed for each spike. Therefore, such kind of coupled common neurons can realize ternary output.

We note that the SpikeGrad algorithm~\citep{thiele2019spikegrad} also requires neurons for ternary output. However, they do not consider how such kind of operation can be implemented with the common neuron models that follow the basic properties of biological neurons, and moreover, they propose another modified neuron model in practice that requires consideration of accumulated spikes for spike generation, which is hardly supported in practice. Differently, our method is with the common neuron model (IF or LIF). This can be suitable for neuromorphic hardware. Also, SpikingYOLO~\citep{kim2020spiking} proposes ternary neurons for the inference of SNNs converted from ANNs without considering the implementation by the common neuron models. Our ternary spiking neuron couples may also provide a way to implement their method on hardware.

\subsection{Solving Implicit Differentiation with Spikes}\label{subsec:solve id}

Based on the coupled neurons in Section~\ref{subsec: coupled neurons}, we can solve implicit differentiation with spikes. For notation simplicity, we directly use Eq.~(\ref{eq.ternary}) as a ternary neuron without detailing coupled neurons below. Our main focus is on solving Eq.~(\ref{eq:linear equation}) with spikes. We first consider the IF neuron model by default and will include the LIF model later. The brief outline for the derivation is: we first derive the update equation of membrane potentials, then we derive the equivalent equation of the rate of spikes with eliminating perturbation, and finally, we prove that the firing rate converges to the solution of Eq.~(\ref{eq:linear equation}).

We first consider the single-layer condition. Let $\bm{\alpha}[T_F]$ denote the average firing rate of these neurons after the forward computation with time steps $T_F$ as an approximate equilibrium state (we treat the forward procedure as the first stage), $\mathbf{g}=\left(\frac{\partial \mathcal{L}}{\partial \bm{\alpha}[T_F]}\right)^\top$ denote the gradient of the loss function on this approximate equilibrium state, and $\mathbf{m}=\sigma'(\bm{\alpha}[T_F]), \mathbf{M}=\text{Diag}(\mathbf{m})$ denote a mask indicator based on the firing condition in the first stage, where $\sigma'(x)=\left\{\begin{aligned}
& 1,\,0<x<1\\
& 0,\,\text{else}\\
\end{aligned}\right.$. We will have another $T_B$ time steps in the second backward stage to calculate implicit differentiation. We set the input to these neurons as $\mathbf{g}$ at all time steps, which can be viewed as input currents~\citep{zhang2020temporal,xiao2021training}. Then along the inverse connections of neurons and with a mask on neurons or weights and an output rescale, the computation of FSNN with ternary neurons is:
\begin{equation}
\begin{aligned}
    \mathbf{u}[t+1] = & \mathbf{u}[t] + \frac{1}{V_{th}-u_{reset}}(\mathbf{M}\mathbf{W})^\top\mathbf{s}[t] + \mathbf{g} - (V_{th}^{b}-u_{reset}^{b})\mathbf{s}[t+1],\\
\end{aligned}
    \label{eq.singlelayer_backward}
\end{equation}
where $V_{th}, u_{reset}$ and $V_{th}^{b}, u_{reset}^{b}$ are the threshold and reset potential during the first and second stage, respectively.
Define the `average firing rate' at this second stage as $\bm{\beta}[t]=\frac{1}{t}\sum_{\tau=1}^t \mathbf{s}[\tau]$, and $\mathbf{u}[0]=\mathbf{0}, \mathbf{s}[0]=\mathbf{0}$, then through summation, we have  (let $V_u=V_{th}-u_{reset}, V_u^b=V_{th}^{b}-u_{reset}^{b}$):
\begin{equation}
\begin{aligned}
    \bm{\beta}[t+1] = & \frac{1}{V_u^b}\left(\frac{t}{t+1}\frac{1}{V_u}(\mathbf{M}\mathbf{W})^\top\bm{\beta}[t]+\mathbf{g}-\frac{\mathbf{u}[t+1]}{t+1}\right).
\end{aligned}
    \label{eq.singlelayer_backward_beta}
\end{equation}
Since there would be at most $t$ spikes during $t$ time steps, $\bm{\beta}$ should be bounded in the range of $[-1,1]$. The membrane potential $\mathbf{u}_i[t]$ will maintain the exceeded terms, i.e. define $\mathbf{v}_i[t]=\left(\frac{t}{t+1}\frac{1}{V_u}(\mathbf{M}\mathbf{W})^\top\bm{\beta}[t]+\mathbf{g}\right)_i$, we can divide $\mathbf{u}_i[t]$ as $\mathbf{u}_i^E[t]+\mathbf{u}_i^B[t]$, where $\mathbf{u}_i^E[t]=\max\left(\mathbf{v}_i[t]-V_{th}^b, 0\right)+\min\left(\mathbf{v}_i[t]+V_{th}^b, 0\right)$ is the exceeded term while $\mathbf{u}_i^B[t]$ is a bounded term~\citep{xiao2021training} which is typically bounded in the range of $[-V_{th}^b, V_{th}^b]$. Then, Eq.~(\ref{eq.singlelayer_backward_beta}) turns into:
\begin{equation}
\begin{aligned}
    \bm{\beta}[t+1] = & \phi\left(\frac{1}{V_u^b}\left(\frac{t}{t+1}\frac{1}{V_u}(\mathbf{M}\mathbf{W})^\top\bm{\beta}[t]+\mathbf{g}\right)\right)-\frac{1}{V_u^b}\frac{\mathbf{u}^B[t+1]}{t+1},
\end{aligned}
    \label{eq.singlelayer_backward_beta_clamp}
\end{equation}
where $\phi(x)=\min(1, \max(-1, x))$. Note that if $\mathbf{g}$ and $(\mathbf{M}\mathbf{W})^\top$ are in an appropriate range, there would be no exceeded term and $\phi$ will not take effect. Indeed we will rescale the loss to control the range of $\mathbf{g}$, as will be indicated in Section~\ref{subsec:error}. With this consideration, we can prove that $\bm{\beta}[t]$ converges to the solution of Eq.~(\ref{eq:linear equation}).
\begin{theorem}\label{thm1}
If there exists $\gamma<1$ such that $\lVert (\mathbf{M}\mathbf{W})^\top \rVert_2 \leq \gamma (V_{th}-u_{reset})(V_{th}^b-u_{reset}^b)$, then the average firing rate $\bm{\beta}[t]$ will converge to an equilibrium point $\bm{\beta}[t]\rightarrow \bm{\beta}^*$. When $V_{th}^b-u_{reset}^b=1$, and there exists $\lambda<1$ such that $\lVert(\mathbf{M}\mathbf{W})^\top\rVert_{\infty}\leq\lambda(V_{th}-u_{reset})$ and $\lVert\mathbf{g}\rVert_{\infty}\leq 1-\lambda$, then $\bm{\beta}^*$ is the solution of Eq.~(\ref{eq:linear equation}).
\end{theorem}
The proof and discussion of assumptions are in~\ref{appsec:proof1}. With Theorem~\ref{thm1}, we can solve Eq.~(\ref{eq:linear equation}) by simulating this second stage of SNN computation to obtain the `firing rate' $\bm{\beta}[T_B]$ as the approximate solution. Plugging this solution to Eq.~(\ref{eq:gradients implicit differentiation}), the gradients can be calculated by: $\nabla_{\mathbf{W}}\mathcal{L}=\frac{1}{V_{th}-u_{reset}}\mathbf{M}\bm{\beta}[T_B]\bm{\alpha}[T_F]^\top, \nabla_{\mathbf{F}}\mathcal{L}=\frac{1}{V_{th}-u_{reset}}\mathbf{M}\bm{\beta}[T_B]\mathbf{\overline{x}}[T_F]^\top,$ and\\ $\nabla_{\mathbf{b}}\mathcal{L}=\frac{1}{V_{th}-u_{reset}}\mathbf{M}\bm{\beta}[T_B]$.

Note that in practice, even if the data distribution is not in a proper range, we can still view $\phi$ as a kind of clipping for improperly large numbers, which is similar to empirical techniques like ``gradient clipping'' to stabilize the training.

Then we consider the extension to the multi-layer condition. Let $\bm{\alpha}^l[T_F]$ denote the average firing rate of neurons in layer $l$ after the forward computation, $\mathbf{g}=\left(\frac{\partial \mathcal{L}}{\partial \bm{\alpha}^N[T_F]}\right)^\top$ denote the gradient of the loss function on the approximate equilibrium state of the last layer, and $\mathbf{m}^l=\sigma'(\bm{\alpha}^l[T_F]), \mathbf{M}^l=\text{Diag}(\mathbf{m}^l)$ denote the mask indicators. Similarly, we will have another $T_B$ time steps in the second stage and set the input to the last layer as $\mathbf{g}$ at all time steps. The computation of FSNN with ternary neurons is calculated as:
\begin{equation}
    \left\{
    \begin{aligned}
        \mathbf{u}^N[t + 1] = &\mathbf{u}^N[t] + \frac{1}{V_u} (\mathbf{M}^1\mathbf{W}^1)^\top\mathbf{s}^1[t] + \mathbf{g}-V_u^b\mathbf{s}^N[t+1],\\
        \mathbf{u}^{l}[t + 1] = &\mathbf{u}^{l}[t] + \frac{1}{V_u}(\mathbf{M}^{l+1}\mathbf{F}^{l+1})^\top\mathbf{s}^{l+1}[t+1]-V_u^b\mathbf{s}^{l}[t+1], (l=N-1,\cdots, 1).\\
    \end{aligned}
    \right.
    \label{eq.multilayer_backward}
\end{equation}
The `average firing rates' $\bm{\beta}^l[t]$ are similarly defined for each layer, and the equivalent form can be similarly derived as:
\begin{equation}
    \left\{
    \begin{aligned}
        \bm{\beta}^N[t+1] =& \phi\left(\frac{1}{V_u^b}\left(\frac{t}{t+1}\frac{1}{V_u}(\mathbf{M}^1\mathbf{W}^1)^\top\bm{\beta}^1[t]+\mathbf{g}\right)\right)-\frac{1}{V_u^b}\frac{{\mathbf{u}^N}^B[t+1]}{t+1},\\
        \bm{\beta}^l[t+1] =& \phi\left(\frac{1}{V_u^b}\left(\frac{1}{V_u}(\mathbf{M}^{l+1}\mathbf{F}^{l+1})^\top\bm{\beta}^{l+1}[t+1]\right)\right)-\frac{1}{V_u^b}\frac{{\mathbf{u}^l}^B[t+1]}{t+1}.\\
    \end{aligned}
    \right.
    \label{eq.multilayer_backward_beta_clamp}
\end{equation}
The convergence of the `firing rate' at the last layer to the solution of Eq.~(\ref{eq:linear equation}) can be similarly derived as Theorem~\ref{thm1}. However, we need to calculate gradients for each parameter as Eq.~(\ref{eq:gradients implicit differentiation}), which is more complex than the single-layer condition. Actually, we can derive that the `firing rates' at each layer converge to equilibrium points, based on which the gradients can be easily calculated with information from the adjacent layers. Theorem~\ref{thm2} gives a formal description.
\begin{theorem}\label{thm2}
If there exists $\gamma<1$ such that $\lVert (\mathbf{M}^1\mathbf{W}^1)^\top\rVert_2\lVert (\mathbf{M}^N\mathbf{F}^N)^\top\rVert_2\cdots \\ \lVert (\mathbf{M}^2\mathbf{F}^2)^\top\rVert_2\leq \gamma {(V_{th}-u_{reset})}^N{(V_{th}^b-u_{reset}^b)}^N$, then the average firing rates $\bm{\beta}^l[t]$ will converge to equilibrium points $\bm{\beta}^l[t]\rightarrow {\bm{\beta}^l}^*$. When $V_{th}^b-u_{reset}^b=1$, and there exists $\lambda<1$ such that $\lVert(\mathbf{M}^1\mathbf{W}^1)^\top\rVert_{\infty}\leq\lambda(V_{th}-u_{reset}), \lVert(\mathbf{M}^l\mathbf{F}^l)^\top\rVert_{\infty}\leq\lambda(V_{th}-u_{reset}), l=2,\cdots,N$ and $\lVert\mathbf{g}\rVert_{\infty}\leq 1-\lambda^N$, then ${\bm{\beta}^N}^*$ is the solution of Eq.~(\ref{eq:linear equation}), and ${\bm{\beta}^l}^*=\left(\frac{\partial h_N({\bm{\alpha}^N}^*)}{\partial h_l({\bm{\alpha}^N}^*)}\right)^{\top}{\bm{\beta}^N}^*, l=N-1,\cdots,1$, where $h_l({\bm{\alpha}^N}^*)=f_l\circ\cdots\circ f_2\left(f_1({\bm{\alpha}^N}^*, \mathbf{x}^*)\right), l=N, \cdots, 1$.
\end{theorem}
The functions $f_l$ are defined in Section~\ref{equilibrium states}. For the proof please refer to~\ref{appsec:proof2}. With Theorem~\ref{thm2}, by putting the solutions into Eq.~(\ref{eq:gradients implicit differentiation}),  the gradients can be calculated by:  $\nabla_{\mathbf{F}^l}\mathcal{L}=\frac{1}{V_{th}-u_{reset}}\mathbf{M}^l\bm{\beta}^l[T_B]\bm{\alpha}^{l-1}[T_F]^\top$ $(l=2,\cdots,N)$, $\nabla_{\mathbf{F}^1}\mathcal{L}=\frac{1}{V_{th}-u_{reset}}\mathbf{M}^1\bm{\beta}^1[T_B]\mathbf{\overline{x}}[T_F]^\top$, $\nabla_{\mathbf{W}^1}\mathcal{L}=\frac{1}{V_{th}-u_{reset}}\mathbf{M}^l\bm{\beta}^l[T_B]\bm{\alpha}^{N}[T_F]^\top$, and $\nabla_{\mathbf{b}^l}\mathcal{L}=\frac{1}{V_{th}-u_{reset}}\mathbf{M}^l\bm{\beta}^l[T_B]$ $(l=2,\cdots,N)$.

Note that the gradient calculation shares an interesting local property, i.e. it is proportional to the firing rates of the two neurons connected by it: $\nabla_{\mathbf{F}^l_{i,j}}\mathcal{L}=\frac{1}{V_{th}-u_{reset}}\mathbf{m}^l_i\bm{\beta}^l_i\bm{\alpha}^{l-1}_j$.
During calculation, since we will have the firing rate of the first stage before the second stage, this calculation can also be carried out by event-based calculation triggered by the spikes in the second stage. So the weight update could be event-driven as well.

Also, note that the theorems still hold if we degrade our feedback models to feedforward ones by setting feedback connections as zero. In this setting, the dynamics and equilibriums degrade to direct functional mappings, and the implicit differentiation degrades to the explicit gradient. We can still approximate gradients with this computation.

In the following, we take $V_{th}^b-u_{reset}^b=1$ by default to fulfill the assumption of theorems (it may take other values if we correspondingly rescale the outputs and we set $1$ for simplicity). Other techniques like dropout can also be included. Please refer to~\ref{appsec:training details} for details.

\paragraph{SPIDE with LIF neuron model}
The above conclusions show that we can leverage the equilibrium states with the IF neuron model to solve implicit differentiation with spikes. As introduced in Section~\ref{equilibrium states}, we can also derive the equilibrium states with the LIF neuron model by considering the weighted average firing rate, and the equilibrium fixed-point equations would have the same form as those of the IF model except for some bounded random error.
Therefore, with the same thought, our SPIDE method can leverage LIF neurons to approximate the solution of implicit differentiation as well. We can replace IF neurons and average firing rates with LIF neurons and weighted average firing rates, and it will gradually approximate the same equilibrium states as in Theorem~\ref{thm1} and Theorem~\ref{thm2} with bounded random error. The derivation is similar to \citet{xiao2021training} and Theorems, and the repetitive details are omitted here. 

\subsection{Reducing Approximation Error}\label{subsec:error}

Section~\ref{subsec:solve id} derives that we can solve implicit differentiation with spikes, as the average firing rate will gradually converge to the solution. In practice, however, we will simulate SNNs for finite time steps, and a smaller number of time steps is better for lower energy consumption. This will introduce approximation error which may hamper training. In this subsection, we theoretically study the approximation error and propose to adjust the reset potential to reduce it. Inspired by the theoretical analysis on quantized gradients~\citep{NEURIPS2020_099fe6b0}, we will analyze the error from the statistical perspective.

For the `average firing rates' $\bm{\beta}^l[t]$ in Eq.~(\ref{eq.singlelayer_backward_beta}) and the multilayer counterparts, the approximation error $e$ to the equilibrium states consists of three independent parts $e_e$, $e_r$ and $e_i$: the first is ${\mathbf{u}^l}^E[t+1]$ that is the exceeded term due to the limitation of spike number, the second is ${\mathbf{u}^l}^B[t+1]$ which can be viewed as a bounded random variable, and the third is the convergence error of the iterative update scheme without ${\mathbf{u}}^l[t+1]$, i.e. let $\mathbf{b}^l[t]$ denote the iterative sequences for solving ${\bm{\beta}^l}^*$ as $\mathbf{b}^l[t+1]=\frac{t}{t+1}\frac{1}{V_{th}-u_{reset}}(\mathbf{M}^{l+1}\mathbf{F}^{l+1})^\top\mathbf{b}^l[t]$, the convergence error is $\lVert \mathbf{b}^l[t]-{\bm{\beta}^l}^*\rVert$. The second part $e_r$ can be again decomposed into two independent components $e_r=e_q+e_s$: $e_q$ is the quantization effect due to the precision of firing rates ($\frac{1}{T}$ for $T$ time steps) if we first assume the same average inputs at all time steps, and $e_s$ is due to the random arrival of spikes rather than the average condition, as there might be unexpected output spikes, e.g. the average input is $0$ and the expected output should be $0$, but two large positive inputs followed by one larger negative input at the last time would generate two positive spikes while only one negative spike. So the error is divided into: $e=e_e+e_q+e_s+e_i$. Since the iterative formulation is certain for $e_i$, we focus on $e_e$, $e_q$ and $e_s$.

Firstly, the error $e_q$ due to the quantization effect is influenced by the input scale and time steps $T_B$. To enable proper input scale and smaller time steps, we rescale the loss function by a factor $s_l$, since the magnitude of gradients with cross-entropy loss is relatively small. We scale the loss to an appropriate range so that information can be propagated by SNNs in smaller time steps, and most signals are in the range of $\phi$ as analyzed in Section~\ref{subsec:solve id}. The base learning rate is scaled by $\frac{1}{s_l}$ correspondingly. This is also adopted by~\citet{thiele2019spikegrad}.

Then given the scale and number of time steps, $e_q$, $e_e$ and $e_s$ can be treated as random variables from statistical perspective, and we view $\bm{\beta}^l[t]$ as stochastic estimators for the equilibrium states with $e_i$. For the stochastic optimization algorithms, the expectation and variance of the gradients are important for convergence and convergence rate~\citep{bottou2010large}, i.e. we hope an unbiased estimation of gradients and smaller estimation variance. As for $e_e$ and $e_s$, they depends on the input data and the expectations are $\mathbb{E}[e_e]=0, \mathbb{E}[e_s]=0$ (the positive and negative parts have the same probability). While for $e_q$, it will depend on our hyperparameters $V_{th}^b$ and $u_{reset}^b$. Since the remaining terms in ${\mathbf{u}^l}^B[t+1]$ caused by the quantization effect is in the range of $[u_{reset}^b, V_{th}^b]$ for positive terms while $[-V_{th}^b, -u_{reset}^b]$ for negative ones, given $V_{th}^b-u_{reset}^b$ and considering the uniform distribution, only when $u_{reset}^b=-V_{th}^b$, $\mathbb{E}[e_q]=0$ for both positive and negative terms. Therefore, we should adjust the reset potential from commonly used 0 to $-V_{th}^b$ for unbiased estimation, as described in Proposition~\ref{pro1}.
\begin{proposition}\label{pro1}
For fixed $V_{th}^b-u_{reset}^b$ and uniformly distributed inputs and $e_q$, only when $u_{reset}^b=-V_{th}^b$, $\bm{\beta}^l[t]$ are unbiased estimators for $\mathbf{b}^l[t]$.
\end{proposition}
Also, taking $u_{reset}^b=-V_{th}^b$ achieves the smallest estimation variance for the quantization effect $e_q$, considering the uniform distribution on $[u_{reset}^b, V_{th}^b]\cup[-V_{th}^b, -u_{reset}^b]$. Since the effects of $e_e$, $e_s$ and $e_i$ are independent of $e_q$ and their variance is certain given inputs, it leads to Proposition~\ref{pro2}.
\begin{proposition}\label{pro2}
Taking $u_{reset}^b=-V_{th}^b$ reduces the variance of estimators $\bm{\beta}^l[t]$.
\end{proposition}
With this analysis, we will take $V_{th}^b=0.5, u_{reset}^b=-0.5$ in the following. For $V_{th}$ and $u_{reset}$ during the first forward stage, we will also take $u_{reset}=-V_{th}$.

\subsection{Details and Training Pipeline}\label{subsec:pipeline}

The original IDE method~\citep{xiao2021training} leverages other training techniques including modified batch normalization (BN) and restriction on weight spectral norm. Since the batch statistical information might be hard to obtain for calculation of biological systems and neuromorphic hardware, and training with the BN operation is hard to be implemented by spike-based calculation, we drop BN in our SPIDE method. The restriction on the weight norm, however, is necessary for the convergence of feedback models, as indicated in theorems. We will adjust it for a more friendly calculation, please refer to~\ref{appsec:training details} for details.

\begin{algorithm}[ht!]\scriptsize
    \caption{Forward procedure of SPIDE training - Stage 1.}
    \hspace*{0.02in} {\bf Input:}
    Network parameters $F^1, b^1,\cdots, F^N, b^N, W^1, W^o, b^o$; Time steps $T_F$; Forward threshold $V_{th}$; Dropout rate $r$; Input data $\{x[t]\}_{t=1}^{T_F}$;\\
    \hspace*{0.02in} {\bf Output:}
    Output of the readout layer $o$.
    \begin{algorithmic}[1]
    \State Initialize $u^i[0]=0 (i=1,2,\cdots,N$), $o=0$
    \State If use dropout, randomly generate dropout masks $D^i (i=1,2,\cdots,N), D^f$ with rate $r$ \quad// $D^i, D^f$ are saved for backward
    \For{$t=1,2,\cdots, T_F$}
        \If{$t==1$}
            \State $u^1[t]=u^1[t-1]+D^1\odot (F^1x[t]+b^1)$
        \Else
            \State $u^1[t]=u^1[t-1]+D^1\odot (F^1x[t]+b^1)+D^f\odot (W^1s^N[t-1])$
        \EndIf
        \State $s^1[t]=H(u^1[t] - V_{th})$
        \State $u^1[t]=u^1[t]-2V_{th}s^1[t]$ \quad// $u_{reset}=-V_{th}$, the same below
        \For{$l=2,3,\cdots,N$}
            \State $u^l[t] = u^l[t-1]+D^l\odot (F^ls^{l-1}[t]+b^l)$
            \State $s^l[t] = H(u^l[t] - V_{th})$
            \State $u^l[t] = u^l[t]-2V_{th}s^l[t]$
        \EndFor
        \State $o = o + W^os^N[t]+b^o$ \quad// $o$ can be accumulated here, or calculated later by $o=W^o\alpha^N+b^o$
    \EndFor
    \State $o = \frac{o}{T}$
    \State $\alpha^i = \frac{\sum_{t=1}^T s^i[t]}{T}, i=1,2,\cdots,N$ \quad// Save for backward, firing rate in Stage 1
    \State $m^i = (\alpha^i>0)\land (\alpha^i<1)$ \quad// Save for backward, mask
    \State If $x$ is not constant, save $x=\frac{\sum_{t=1}^Tx[t]}{T}$ for backward
    \State \textbf{Return} $o$
    \end{algorithmic}
    \label{algorithm: forward}
\end{algorithm}

\begin{algorithm}[ht!]\scriptsize
    \caption{Backward procedure of SPIDE training - Stage 2.}
    \hspace*{0.02in} {\bf Input:}
    Network parameters $F^1, b^1,\cdots, F^N, b^N, W^1, W^o, b^o$; Forward output $o$; Label $y$; Time steps $T_B$; Forward threshold $V_{th}$; Backward threshold $V_{th}^b=0.5$; Other hyperparameters and saved variables;\\
    \hspace*{0.02in} {\bf Output:}
    Trained network parameters $F^1, b^1,\cdots, F^N, b^N, W^1, W^o, b^o$.
    \begin{algorithmic}[1]
    \State Calculate $g=\frac{\partial L(o, y)}{\partial o}$ \quad// for CE loss, $\frac{\partial L(o, y)}{\partial o}=\text{softmax}(o)-y$, in practice we will scale the loss by a factor $s_l$, then $\frac{\partial L(o, y)}{\partial o}=s_l\left(\text{softmax}(o)-y\right)$
    \State Initialize $u^i[0]=0, i=1,2,\cdots,N$
    \For{$t=1,2,\cdots, T_B$}
        \If{$t==1$}
            \State $u^N[t]=u^N[t-1]+{W^o}^\top g$
        \Else
            \State $u^N[t]=u^N[t-1]+{W^o}^\top g+\frac{1}{2V_{th}}{W^1}^\top(D^f\odot m^1\odot s^1[t-1])$ \quad// $m^i$ is the saved mask in Stage 1
        \EndIf
        \State $s^N[t]=T(u^N[t], 0.5)$ \quad// realized by two coupled neurons
        \State $u^N[t]=u^N[t]-s^N[t]$ \quad// realized by two coupled neurons
        \For{$l=N-1,N-2,\cdots,1$}
            \State $u^l[t] = u^l[t-1]+\frac{1}{2V_{th}}{F^l}^\top(D^l\odot m^{l+1}\odot s^{l+1}[t])$ \quad// $m^i$ is the saved mask in Stage 1
            \State $s^l[t] = T(u^l[t], 0.5)$ \quad// realized by two coupled neurons
            \State $u^l[t] = u^l[t]-s^l[t]$ \quad// realized by two coupled neurons
        \EndFor
    \EndFor
    \State $\beta^i = \frac{\sum_{t=1}^T s^i[t]}{T}, i=1,2,\cdots,N$ \quad// ``firing rate'' in Stage 2
    \State Calculate gradients: \quad// Note that the below calculation can be realized by event-driven accumulation based on the above $s^i[t]$ considering the definition of $\beta^i$
    \State \hspace*{0.1in} (1) $\nabla_{F^1} \mathcal{L} = \frac{1}{2V_{th}}(m^1\odot \beta^1)x^\top$ \quad// $m^i, x$ are the saved mask and average input in Stage 1
    \State \hspace*{0.1in} (2) $\nabla_{F^i} \mathcal{L} = \frac{1}{2V_{th}}(m^i\odot \beta^i){\alpha^{i-1}}^\top, i=2,3,\cdots,N$ \quad// $m^i, \alpha^i$ are the saved mask and firing rate in Stage 1
    \State \hspace*{0.1in} (3) $\nabla_{b^i} \mathcal{L} = \frac{1}{2V_{th}}(m^i\odot \beta^i), i=1,2,\cdots,N$
    \State \hspace*{0.1in} (4) $\nabla_{W^1} \mathcal{L} = \frac{1}{2V_{th}}(m^1\odot \beta^1){\alpha^{N}}^\top$ \quad// $m^i, \alpha^i$ are the saved mask and firing rate in Stage 1
    \State \hspace*{0.1in} (5) $\nabla_{W^o} \mathcal{L} = \alpha^{N}\left(\frac{\partial L(o, y)}{\partial o}\right)^\top$ \quad// $\alpha^i$ is the saved firing rate in Stage 1
    \State \hspace*{0.1in} (6) $\nabla_{b^o} \mathcal{L} = \left(\frac{\partial L(o, y)}{\partial o}\right)^\top$
    \State Update $F^1, b^1,\cdots, F^N, b^N, W^1, W^o, b^o$ based on the gradient-based optimizer \quad// SGD learning rate $\eta$ + momentum $\alpha$ \& weight decay $\mu$, the base learning rate is scaled by the factor $s_l$ of the loss, i.e. $\eta=\frac{\eta}{s_l}$
    \State \hspace*{0.1in} (1) Update the momentum $M_{\theta}=\alpha*M_{\theta}+(1-\alpha)*\nabla_{\theta} \mathcal{L}, \theta\in\{F^i, b^i, W^1, W^o, b^o\}$
    \State \hspace*{0.1in} (2) Update parameters $\theta=(1-\mu)*\theta+\eta*M_{\theta}, \theta\in\{F^i, b^i, W^1, W^o, b^o\}$
    \State \hspace*{0.1in} (3) Restrict the norm of $W^1$
    \State \textbf{Return} $F^1, b^1,\cdots, F^N, b^N, W^1, W^o, b^o$
    \end{algorithmic}
    \label{algorithm: backward}
\end{algorithm}

We summarize our training pipeline as follows. There are two stages for the forward and backward procedures respectively. In the first stage, SNNs receive inputs and perform the calculation as Eq.~(\ref{eq.snn},\ref{eq.singlelayer},\ref{eq.multilayer}) for $T_F$ time steps, after which we get the output from the readout layer, and save the average inputs as well as the average firing rates and masks of each layer for the second stage. In the second stage, the last layer of SNNs will receive gradients for outputs and perform calculation along the inverse connections as Eq.~(\ref{eq.ternary},\ref{eq.singlelayer_backward},\ref{eq.multilayer_backward}) for $T_B$ time steps, after which we get the `average firing rates' of each layer. With firing rates from two stages, the gradients for parameters can be calculated as in Section~\ref{subsec:solve id} and then the first-order optimization algorithm is applied to update the parameters. We provide detailed pseudocodes for both stages in Algorithm~\ref{algorithm: forward} and Algorithm~\ref{algorithm: backward}, respectively. Figure~\ref{fig: illustration} also illustrates the overall process.

\section{Experiments}\label{sec:experiments}

In this section, we conduct experiments to demonstrate the effectiveness of our method and the great potential for energy-efficient training. We simulate the computation on common computational units. Please refer to~\ref{appsec:training details} for implementation details and descriptions.

\begin{table} [ht]
	\centering
	\small
	\captionof{table}{Evaluation of training with different time steps in the backward stage. Training is on CIFAR-10 with AlexNet-F structure and $T_F=30$. Results are based on 3 runs of experiments.}
	\begin{tabular}{c|c}
		\toprule[1pt]
		$T_B$ & Mean$\pm$Std (Best)\\
		\midrule[0.5pt]
		50 & 88.41\%$\pm$0.48\% (89.07\%)\\
		100 & 89.17\%$\pm$0.14\% (89.35\%)\\
		250 & 89.61\%$\pm$0.11\% (89.70\%)\\
		500 & 89.57\%$\pm$0.08\% (89.67\%)\\
		\bottomrule[1pt]
	\end{tabular}
	\label{analysis time steps backward}
\end{table}

\paragraph{Effectiveness with a small number of backward time steps} As shown in Table~\ref{analysis time steps backward}, we can train high-performance models with low latency ($T_F=30$) in a small number of backward time steps during training (e.g. $T_B=50$), which indicates the low latency and high energy efficiency. Note that conventional ANN-SNN methods require hundreds to thousands of time steps just for satisfactory inference performance, and recent progress of the conversion methods and direct training methods show that relatively small time steps are enough for inference, while we are the first to demonstrate that even training of SNNs can be carried out with spikes in a very small number of time steps. This is due to our analysis and improvement to reduce the approximation error, as illustrated in the following ablation study.

\begin{figure}
    \hspace{-7em}
    \begin{subfigure} {0.7\linewidth}
    \includegraphics[width=\linewidth]{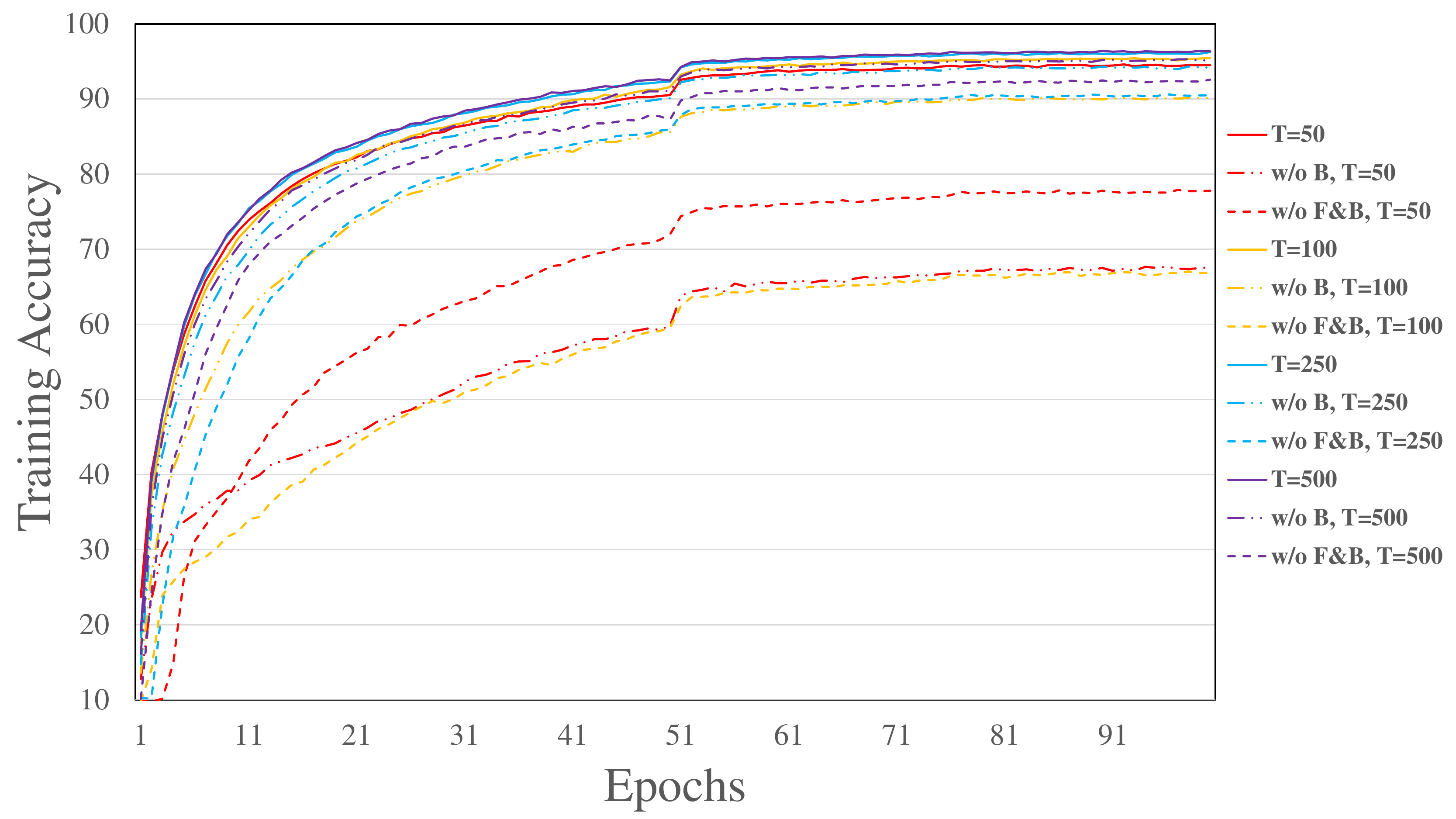}
    \end{subfigure}
    \begin{subfigure}{0.7\linewidth}
    \includegraphics[width=\linewidth]{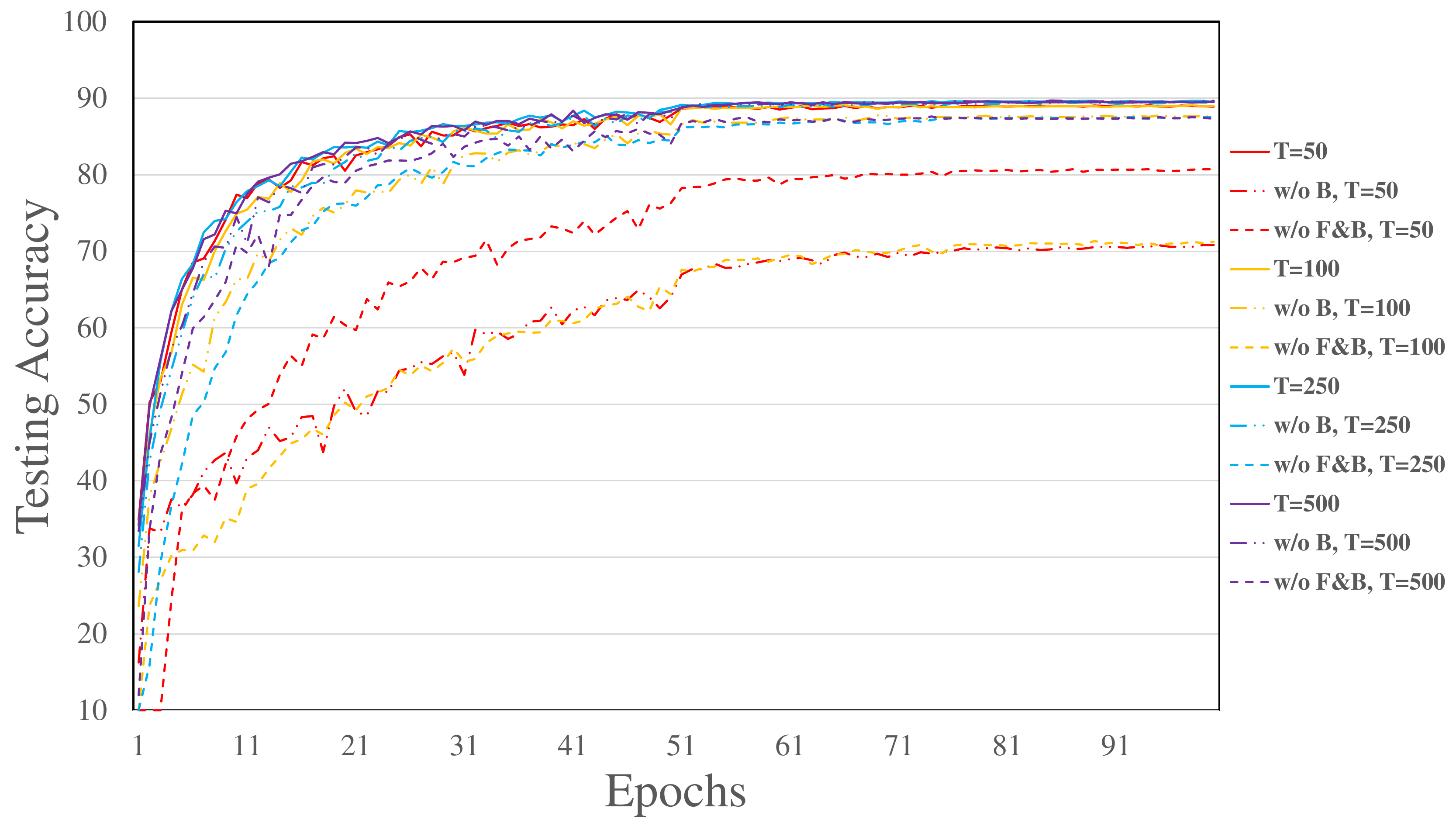}
    \end{subfigure}
    \caption{Comparison of training and testing curves under different settings and backward time steps.}
    \label{fig: ablation}
\end{figure}

\paragraph{Ablation study of reducing the approximation error}
We conduct ablation study on our improvement to reduce the approximation error by setting the reset potential as negative threshold. To formulate equivalent equilibrium states, we take the same $V_{th}-u_{reset}=V_u$ and the same $V_{th}^b-u_{reset}^b=V_u^b$, and we consider the following settings: (1) both forward and backward stages apply our improvement, i.e. $u_{reset}=-V_{th}, u_{reset}^b=-V_{th}^b$; (2) remove the improvement on the backward stage, i.e. $V_{th}^b=V_u^b, u_{reset}^b=0$; (3) remove the improvement on both forward and backward stages, i.e. $V_{th}=V_u, u_{reset}=0$ and $V_{th}^b=V_u^b, u_{reset}^b=0$. The latter two setting are denoted by ``w/o B'' and ``w/o F\&B'' respectively.
The models are trained on CIFAR-10 with AlexNet-F structure and 30 forward time steps. The training and testing curves under different settings and backward time steps are illustrated in Figure~\ref{fig: ablation}. It demonstrates that without our improvement, the training can not perform well within a small number of backward time steps, probably due to the bias and large variance of the estimated gradients. When the backward time steps are large, the performance gap is reduced since the bias of estimation is reduced. It shows the superiority of our improvement in training SNNs within a small number of backward time steps.

\paragraph{Firing rate statistics and potential of energy efficiency}
\begin{figure}[ht]
	\centering
	\small
	\captionof{figure}{Average firing rates for forward and backward stages during training. `A' is AlexNet-F, `C' is CIFARNet-F, and $T$ is time steps for the backward stage.}
	\includegraphics[width=\textwidth]{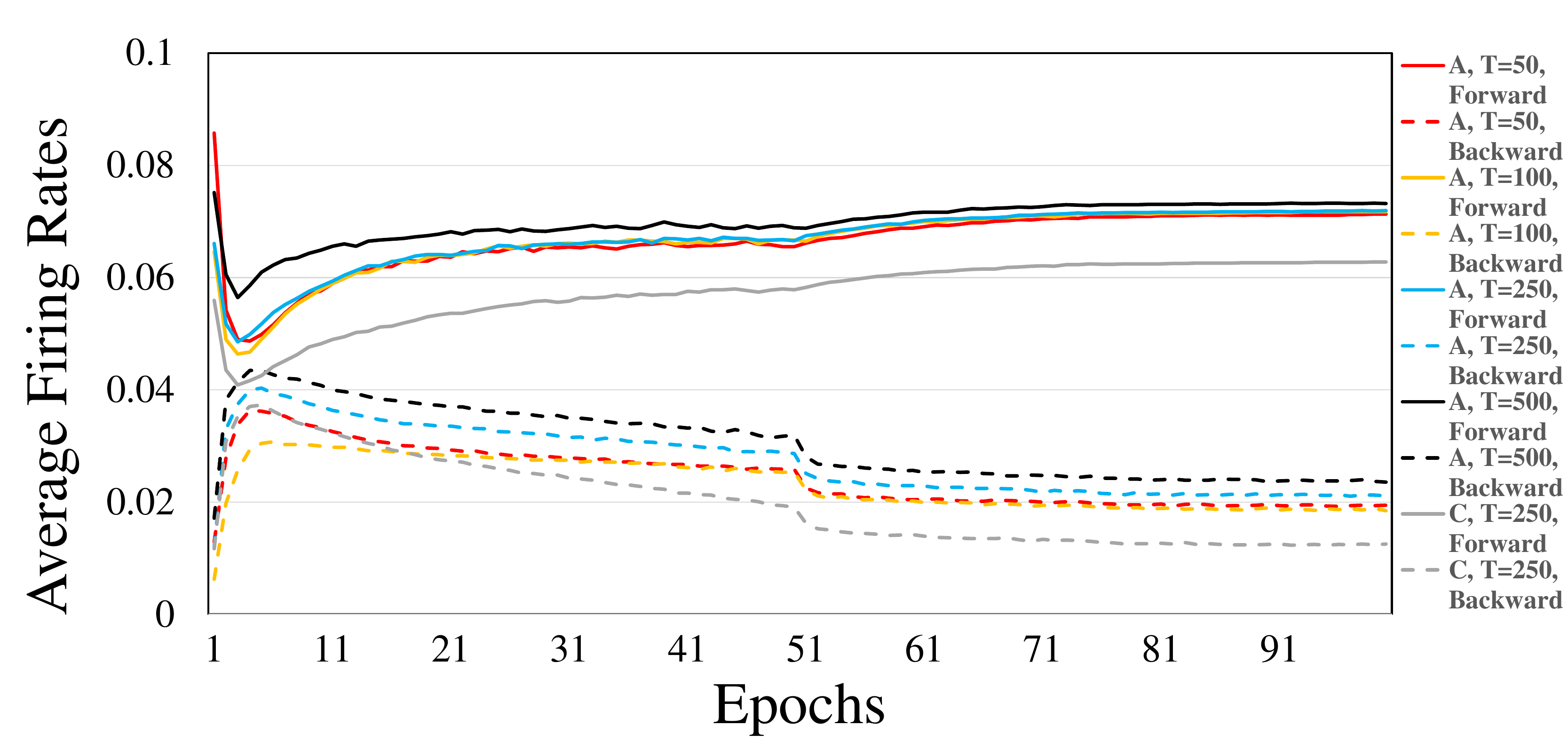}
	\label{firing rate}
\end{figure}
\begin{table} [ht]
	\centering
	\small
	\captionof{table}{Theoretical estimation of energy costs of the backward stage of training with $T_F=30, T_B=50$.}
	\begin{tabular}{c|c|c|c}
		\toprule[1pt]
		Method & OpNum & OpEnergy & Cost\\
		\midrule[0.5pt]
		SPIDE (ours) & $\approx0.03\times 50$ (FiringRate$\times$ $T_B$) & 0.9 (AC) & $1\times$\\
		STBP & 30 ($T_F$) & 4.6 (MAC) & $102\times$\\
		\bottomrule[1pt]
	\end{tabular}
	\label{estimate energy}
\end{table}
\begin{table} [h!]
	\centering
	\small
	\captionof{table}{Theoretical estimation of energy costs of the forward stage with $T_F=30$.}
	\begin{tabular}{c|c|c|c}
		\toprule[1pt]
		Method & OpNum & OpEnergy & Cost\\
		\midrule[0.5pt]
		Spike-based & $\approx0.07\times 30$ (FiringRate$\times$ $T_F$) & 0.9 (AC) & $1\times$\\
		Not spike-based & 30 ($T_F$) & 4.6 (MAC) & $73\times$\\
		\bottomrule[1pt]
	\end{tabular}
	\label{estimate energy forward}
\end{table}
Since the energy consumption of event-driven SNNs is proportional to the number of spikes, we present the average firing rates for forward and backward stages (for backward, both positive and negative spikes are considered as firing) in Figure~\ref{firing rate}.
It shows the firing sparsity of our method, and spikes are sparser in the backward stage with around only $3\%$. Combined with the small number of time steps, this demonstrates the great potential for the energy-efficient training of SNNs with spike-based computation. We theoretically estimate and compare the energy costs for the operations of neurons of our method and the representative STBP method\footnote{The energy costs of the IDE method are hard to estimate because IDE leverages root-finding methods (e.g. Broyden's method) to solve implicit differentiation, which may involve many complex operations and the iteration number may depend on the convergence speed. If we consider the fixed-point update scheme with a fixed iteration number $N$ as the root-finding method in the IDE, its operation number is about the same as STBP with $N$ time steps to backpropagate. As the iteration threshold of IDE is usually taken as 30 (i.e. the maximum iteration number for root-finding methods if the update does not converge before it), the energy estimation of STBP with 30 time steps can also be an effective surrogate result for IDE.}. We mainly focus on the energy of the backward stage which is made spike-based and only requires accumulation (AC) operations by SPIDE while requiring multiply and accumulate (MAC) operations by STBP. Our operation number is estimated as the firing rate multiplied by backward time steps $T_B$, and that of STBP is the forward time steps $T_F$ as STBP will repeat the computation for $T_F$ times to backpropagate through time. According to the 45nm CMOS processor, the energy for 32bit FP MAC operation is 4.6 pJ, and for AC operation is 0.9 pJ. Therefore, as shown in Table~\ref{estimate energy}, when $T_F=30, T_B=50$, our method could achieve approximately $102\times$ reduction in energy cost\footnote{Note that the realization of ternary spiking neuron couples may require twice the synaptic operations under some cases (see Section 4.1), so the energy cost reduction may be halved. Despite this, we could still achieve approximately $50\times$ reduction in energy cost.}. Additionally, if we consider that the forward computation of STBP also has to be simulated by, e.g. GPU, to be compatible with the backward stage and the computation is not spike-based, while ours may be deployed with neuromorphic hardware for spike-based computation, then the forward stage during training can also reduce the energy by around $73\times$, as shown in Table~\ref{estimate energy forward}. Note that even for STBP with fewer forward time steps, e.g. 12 or 6, the energy costs of SPIDE for the backward stage are still about $40\times$ or $20\times$ less than STBP, and our result in Table~\ref{classification performance cifar10} will show that SPIDE is also effective with smaller $T_F$.

\begin{table} [ht]
	\scriptsize
	\tabcolsep=1mm
	\caption{Performance on MNIST with 3 runs of experiments. ``N.A.'' means the method is not spike-based.}
	\hspace{-6em}
	\begin{tabular}{ccccccc}
	    \multicolumn{7}{c}{\textbf{MNIST}}\\
		\toprule[1pt]
		Method & Network structure & $T_F$ & $T_B$ & Mean$\pm$Std (Best) & Neurons & Params \\
		\midrule[0.5pt]
		BP~\citep{lee2016training} & 20C5-P2-50C5-P2-200 & $>$200 & N.A. & (99.31\%) & 33K & 518K\\
		STBP~\citep{wu2018spatio} & 15C5-P2-40C5-P2-300 & 30 & N.A. & (99.42\%) & 26K & 607K\\
		IDE~\citep{xiao2021training} & 64C5 (F64C5) & 30 & N.A. & 99.53\%$\pm$0.04\% (99.59\%) & 13K & 229K\\
		\midrule[0.5pt]
		SpikeGrad~\citep{thiele2019spikegrad} & 15C5-P2-40C5-P2-300 & Unknown & Unknown & 99.38\%$\pm$0.06\% (99.52\%) & 26K & 607K\\
		\textbf{SPIDE (ours)} & 64C5s-64C5s-64C5 (F64C3u) & 30 & 100 & 99.34\%$\pm$0.02\% (99.37\%) & 20K & 275K\\
		\textbf{SPIDE (ours, degraded)} & 15C5-P2-40C5-P2-300 & 30 & 100 & 99.44\%$\pm$0.02\% (99.47\%) & 26K & 607K\\
		\bottomrule[1pt]
	\end{tabular}
	\label{classification performance mnist}
\end{table}
\begin{table} [ht]
	\scriptsize
	\tabcolsep=1mm
	\caption{Performance on CIFAR-10 with 3 runs of experiments. ``N.A.'' means the method is not spike-based.}
	\hspace{-6em}
	\begin{tabular}{ccccccc}
	    \multicolumn{7}{c}{\textbf{CIFAR-10}}\\
		\toprule[1pt]
		Method & Network structure & $T_F$ & $T_B$ & Mean$\pm$Std (Best) & Neurons & Params \\
		\midrule[0.5pt]
		ANN-SNN~\cite{sengupta2019going} & VGG-16 & 2500 & N.A. & (91.55\%) & 311K & 15M\\
		ANN-SNN~\citep{deng2021optimal} & CIFARNet & 400-600 & N.A. & (90.61\%) & 726K & 45M\\
		STBP~\citep{wu2019direct} & AlexNet & 12 & N.A. & (85.24\%) & 595K & 21M\\
		STBP (w/o NeuNorm)~\citep{wu2019direct} & CIFARNet & 12 & N.A. & (89.83\%) & 726K & 45M\\
		STBP~\citep{xiao2021training} & AlexNet-F & 30 & N.A. & (87.18\%) & 159K & 3.7M\\
		IDE~\citep{xiao2021training} & AlexNet-F & 30 & N.A. & 91.74\%$\pm$0.09\% (91.92\%) & 159K & 3.7M\\
		IDE~\citep{xiao2021training} & CIFARNet-F & 30 & N.A. & 92.08\%$\pm$0.14\% (92.23\%) & 232K & 11.8M\\
		\midrule[0.5pt]
		SpikeGrad~\citep{thiele2019spikegrad} & CIFARNet & Unknown & Unknown & 89.49\%$\pm$0.28\% (89.99\%) & 726K & 45M\\
		\textbf{SPIDE (ours)} & AlexNet-F & 12 & 250 & 89.11\%$\pm$0.29\% (89.43\%) & 159K & 3.7M\\
		\textbf{SPIDE (ours)} & AlexNet-F & 30 & 250 & 89.61\%$\pm$0.11\% (89.70\%) & 159K & 3.7M\\
		\textbf{SPIDE (ours)} & CIFARNet-F & 30 & 250 & 89.94\%$\pm$0.17\% (90.13\%) & 232K & 11.8M\\
		\bottomrule[1pt]
	\end{tabular}
	\label{classification performance cifar10}
\end{table}

\paragraph{Competitive performance on common datasets}
We evaluate the performance of our method on static datasets MNIST~\citep{lecun1998gradient}, CIFAR-10, and CIFAR-100~\citep{krizhevsky2009learning}, as well as the neuromorphic dataset CIFAR10-DVS~\citep{li2017cifar10}. We compare our method to several ANN-SNN methods~\citep{sengupta2019going,deng2021optimal}, direct SNN training methods~\citep{wu2018spatio,xiao2021training}, and SpikeGrad~\citep{thiele2019spikegrad} with similar network structures.
As shown in Table~\ref{classification performance mnist} and Table~\ref{classification performance cifar10}, we can train both feedforward and feedback SNN models with a small number of time steps and our trained models achieve competitive results on MNIST and CIFAR-10. Compared with SpikeGrad~\citep{thiele2019spikegrad}, we can use fewer neurons and parameters due to flexible network structure choices, and a small number of time steps while they do not report this important feature. Besides, we use common neuron models while they require special neuron models that are hardly supported, as indicated in Section~\ref{subsec: coupled neurons}.
\begin{figure}
    \centering
    \includegraphics[width=\textwidth]{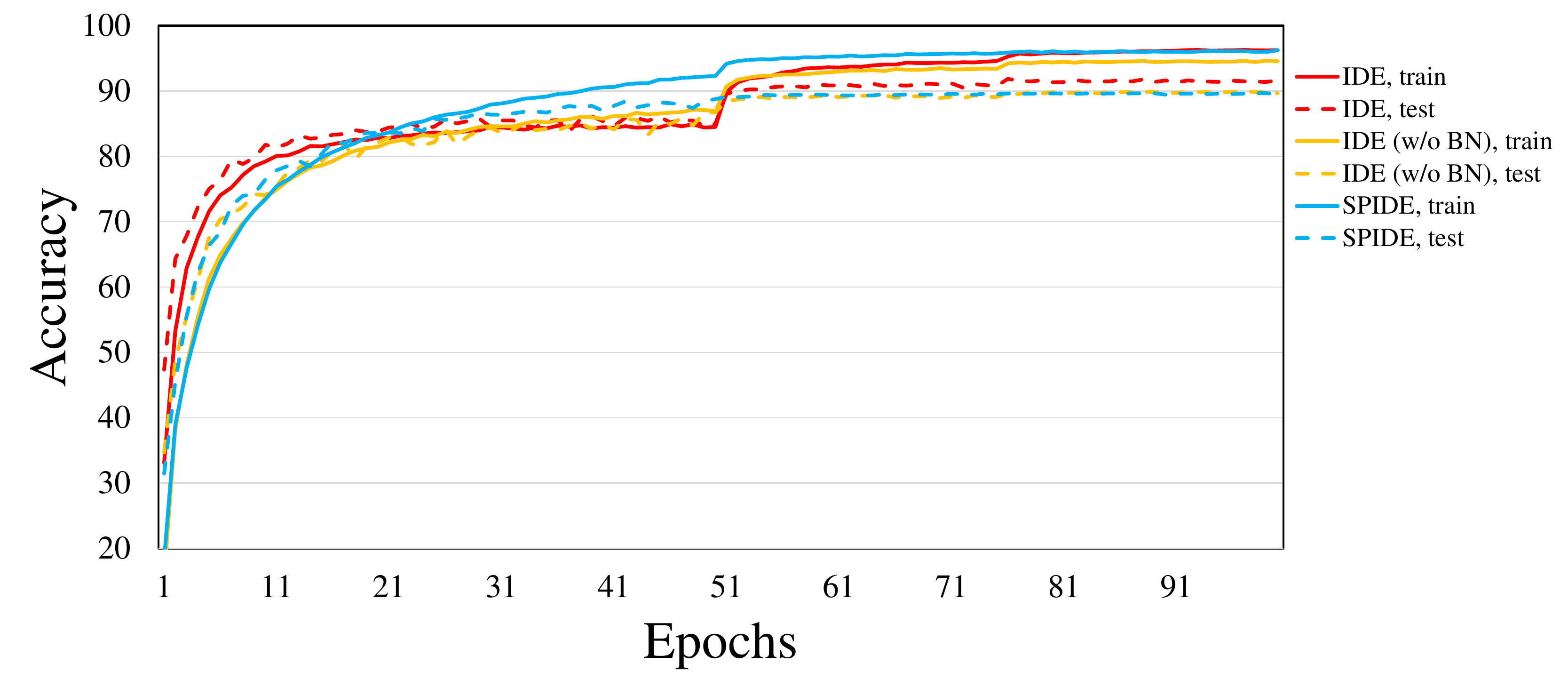}
    \caption{Comparison of training and testing curves between IDE and SPIDE on CIFAR-10 with AlexNet-F structure and $T_F=30$.}
    \label{fig: comparison-ide}
\end{figure}
Compared with the original IDE method~\citep{xiao2021training}, our generalization performance is poorer as we discard the BN component. As shown in Figure~\ref{fig: comparison-ide}, the SPIDE method can achieve the same training accuracy as the IDE method, while the generalization performance is poorer. Since the hyperparameters are the same for experiments, except that we drop the modified BN component (as explained in Section~\ref{subsec:pipeline}), the performance gap may be caused by the implicit regularization effect of BN. We try to drop the modified BN component for IDE and the results in Figure~\ref{fig: comparison-ide} show that it achieves similar performance (89.93\%) as SPIDE. Therefore, the optimization ability of the SPIDE method should be similar to that of IDE, while future work could investigate techniques that are similar to BN but more friendly to the spike-based computation of SNNs to further improve the performance.

\begin{table} [ht]
	\scriptsize
	\tabcolsep=1mm
	\caption{Performance on CIFAR-100 with 3 runs of experiments. ``N.A.'' means the method is not spike-based.}
	\hspace{-6em}
	\begin{tabular}{cccccccc}
		\toprule[1pt]
		Method & Network structure & BN & $T_F$ & $T_B$ & Mean$\pm$Std (Best) & Neurons & Params \\
		\midrule[0.5pt]
		BP~\citep{thiele2019spikegrad} & CIFARNet & $\times$ & Unknown & N.A. & (64.69\%) & 726K & 45M\\
		IDE~\citep{xiao2021training} & CIFARNet-F & \checkmark & 30 & N.A & 71.56\%$\pm$0.31\% (72.10\%) & 232K & 14.8M\\
		\midrule[0.5pt]
		SpikeGrad~\citep{thiele2019spikegrad} & CIFARNet & $\times$ & Unknown & Unknown & (64.40\%) & 726K & 45M\\
		\textbf{SPIDE (ours)} & CIFARNet-F & $\times$ & 30 & 100 & 63.57\%$\pm$0.30\%(63.91\%) & 232K & 14.8M\\
		\textbf{SPIDE (ours)} & CIFARNet-F & $\times$ & 30 & 250 & 64.00\%$\pm$0.11\%(64.07\%) & 232K & 14.8M\\
		\bottomrule[1pt]
	\end{tabular}
	\label{cifar100}
\end{table}

\begin{table} [ht]
	\centering
	\scriptsize
	\tabcolsep=1mm
	\caption{Performance on CIFAR10-DVS.}
	\begin{tabular}{cccccccc}
		\toprule[1pt]
		Method & Model & $T_F$ & $T_B$ & Accuracy \\
		\midrule[0.5pt]
		Gabor-SNN~\citep{sironi2018hats} & Gabor-SNN & N.A & N.A & 24.5\%\\
		HATS~\citep{sironi2018hats} & HATS & N.A & N.A & 52.4\%\\
		STBP~\citep{wu2019direct} & Spiking CNN (LIF, w/o NeuNorm) & 40 & N.A & 58.1\%\\
		STBP~\citep{wu2019direct} & Spiking CNN (LIF, w/ NeuNorm) & 40 & N.A & 60.5\%\\
		Tandem Learning~\citep{wu2021tandem} & Spiking CNN (IF) & 20 & N.A & 58.65\%\\
		ASF-BP~\citep{wu2021training} & Spiking CNN (IF) & Unknown & N.A & 62.5\%\\
		\midrule[0.5pt]
		\textbf{SPIDE (ours)} & Spiking CNN (IF) & 30 & 250 & 60.7\%\\
		\bottomrule[1pt]
	\end{tabular}
	\label{cifar10-dvs}
\end{table}

The results on CIFAR-100 are shown in Table~\ref{cifar100} and our model can achieve 64.07\% accuracy. Compared with IDE, the performance is poorer, and the main reason is probably again the absence of BN which could be important for alleviating overfitting on CIFAR-100 with a relatively small number of images per class. The training accuracy of SPIDE is similar to IDE (around 93\% v.s. around 94\%) while the generalization performance is poorer. Despite this, the performance of our model is competitive for networks without BN and our model is with fewer neurons and parameters and a small number of time steps. Compared with SpikeGrad~\citep{thiele2019spikegrad}, we can use fewer neurons and parameters due to flexible network structure choices, and we leverage common neuron models while they do not. Future work could investigate more suitable structures and more friendly techniques to further improve the performance.

The results on CIFAR10-DVS are shown in Table~\ref{cifar10-dvs}, and our model can achieve 60.7\% accuracy. It is competitive among results of common SNN models, demonstrating the effectiveness of our method.

The above results show the effectiveness of our method even with the constraint of purely spike-based training. We note that there are some recent works that achieve higher state-of-the-art performance~\citep{zheng2020going,Fang_2021_ICCV,li2021differentiable,fang2021deep,deng2021temporal}. However, their target is different from ours which aims at training SNNs with purely spike-based computation as introduced in Section~\ref{introduction}, and they leverage many other techniques such as batch normalization along the temporal dimension or learnable membrane time constant. We do not aim at outperforming the state-of-the-art results but demonstrate that a competitive performance can be achieved even with our constraints of purely spike-based training with common neuron models. And our future work could seek techniques friendly to neuromorphic computation to further improve the performance.

As a preliminary attempt, we provide the result of applying the scaled weight standardization (sWS) technique which is shown as a powerful method to replace BN in ResNets~\citep{brock2021characterizing,brock2021high} and SNNs~\citep{xiao2022online}. Particularly, sWS standardizes weights instead of activations by $\hat{\mathbf{W}}_{i,j}=\gamma \cdot \frac{\mathbf{W}_{i,j}-\mu_{\mathbf{W}_{i,\cdot}}}{\sigma_{\mathbf{W}_{i,\cdot}}\sqrt{N}}$, where $\mu_{\mathbf{W}_{i,\cdot}}$ and $\sigma_{\mathbf{W}_{i,\cdot}}$ are the mean and variance calculated along the input dimension, and the scale $\gamma$ is determined by analyzing the signal propagation with different activation functions (typically taken as $\gamma=\frac{\sqrt{2}}{\sqrt{1-\frac{1}{\pi}}}$). \citet{brock2021characterizing} show that normalization-free ResNets with the sWS technique can achieve a similar performance as common ResNets with BN. We apply this technique to our experiment on CIFAR-10 and the performance of the AlexNet-F structure improves from 89.61\% to 90.37\% and the performance of the CIFARNet-F structure improves from 89.94\% to 90.85\%. It shows that we can effectively leverage other techniques to improve performance.

\begin{table} [ht]
	\centering
	\small
	\tabcolsep=1mm
	\caption{Performance of LIF neurons on MNIST, CIFAR-10, and CIFAR-100. Results are based on 3 runs of experiments.}
	\begin{tabular}{c|cccccccc}
		\toprule[1pt]
		Dataset & Forward Model & Backward Model & $T_F$ & $T_B$ & Mean$\pm$Std (Best) \\
		\midrule[0.5pt]
		\multirow{3}*{MNIST} & IF & IF & 30 & 100 & 99.34\%$\pm$0.02\% (99.37\%)\\
		& LIF & IF & 30 & 100 & 99.32\%$\pm$0.04\% (99.37\%)\\
		& LIF & LIF & 30 & 100 & 99.34\%$\pm$0.05\% (99.39\%)\\
        \hline
        \multirow{3}*{CIFAR-10} & IF & IF & 30 & 250 & 89.94\%$\pm$0.17\% (90.13\%)\\
		& LIF & IF & 30 & 250 & 89.66\%$\pm$0.12\% (89.78\%)\\
		& LIF & LIF & 30 & 250 & 89.54\%$\pm$0.14\% (89.72\%)\\
        \hline
        \multirow{3}*{CIFAR-100} & IF & IF & 30 & 250 & 64.00\%$\pm$0.11\% (64.07\%)\\
		& LIF & IF & 30 & 250 & 64.04\%$\pm$0.03\% (64.06\%)\\
		& LIF & LIF & 30 & 250 & 63.81\%$\pm$0.19\% (63.97\%)\\
		\bottomrule[1pt]
	\end{tabular}
	\label{lif results}
\end{table}

\paragraph{Results with LIF neuron model}
As introduced in Section~\ref{subsec:solve id}, our SPIDE method is also applicable to the LIF neuron model. We conduct experiments on MNIST, CIFAR-10, and CIFAR-100 to verify the effectiveness of SPIDE with LIF neurons. Following \citep{xiao2021training}, the leaky term for LIF neurons is set as $\lambda=0.95$ for MNIST and $\lambda=0.99$ for CIFAR-10 and CIFAR-100. The network structure for CIFAR-10 and CIFAR-100 is taken as CIFARNet-F, and other details are the same as the experiment for IF neurons. The results are shown in Table~\ref{lif results}. It demonstrates that SPIDE is also effective for LIF neurons.

\section{Conclusion}

In this work, we propose the SPIDE method that generalizes the IDE method to enable purely spike-based training of SNNs with common neuron models and flexible network structures.
We prove that the implicit differentiation can be solved with spikes by our coupled neurons. We also analyze the approximation error due to finite time steps and propose to adjust the reset potential of SNNs. Experiments show that we can achieve competitive performance with a small number of training time steps and sparse spikes, which demonstrates the great potential of our method for energy-efficient training of SNNs with spike-based computation.
As for the limitations, SPIDE and IDE mainly focus on the condition that inputs are convergent in the context of average accumulated signals, e.g. image classification with static or neuromorphic inputs. For long-term sequential data such as speech, we may define the final average inputs as the convergent inputs and directly apply our method, but the error flow throughout time is not carefully handled. An interesting future work is to generalize the methodology to time-varying inputs with more careful consideration of error flows and equilibriums, e.g. with certain time windows.

\section*{Acknowledgements}
Z. Lin was supported by the major key project of PCL (No. PCL2021A12), the NSF China (No. 62276004), and Project 2020BD006 supported by PKU-Baidu Fund. Yisen Wang is partially supported by the National Natural Science Foundation of China under Grant 62006153, and Project 2020BD006 supported by PKU-Baidu Fund.

\appendix

\section{Proof of Theorem~\ref{thm1}}\label{appsec:proof1}

\begin{proof}

We first prove the convergence of $\bm{\beta}[t]$.
Let $V_u$ and $V_u^b$ denote $V_{th}-u_{reset}$ and $V_{th}^b-u_{reset}^b$ respectively. Consider $\lVert \bm{\beta}[t+1]-\bm{\beta}[t]\rVert$, it satisfies:
\begin{align}
&\left\lVert \bm{\beta}[t+1]-\bm{\beta}[t]\right\rVert\notag\\
=\quad&\left\lVert\, \left(\phi\left(\frac{1}{V_u^b}\left(\frac{t}{t+1}\frac{1}{V_u}(\mathbf{M}\mathbf{W})^\top\bm{\beta}[t]+\mathbf{g}\right)\right)-\frac{1}{V_u^b}\frac{\mathbf{u}^B[t+1]}{t+1}\right)\right.\notag\\
&\left.- \left(\phi\left(\frac{1}{V_u^b}\left(\frac{t-1}{t}\frac{1}{V_u}(\mathbf{M}\mathbf{W})^\top\bm{\beta}[t-1]+\mathbf{g}\right)\right)-\frac{1}{V_u^b}\frac{\mathbf{u}^B[t]}{t} \right)\right\rVert\notag\\
\leq\quad&\left\lVert\, \phi\left(\frac{1}{V_{u}^b}\left(\frac{1}{V_u}(\mathbf{M}\mathbf{W})^\top\bm{\beta}[t]+\mathbf{g}\right)\right) - \phi\left(\frac{1}{V_{u}^b}\left(\frac{1}{V_u}(\mathbf{M}\mathbf{W})^\top\bm{\beta}[t-1]+\mathbf{g}\right)\right) \right\rVert\notag\\
&+ \left\lVert\, \phi\left(\frac{1}{V_u^b}\left(\frac{t}{t+1}\frac{1}{V_u}(\mathbf{M}\mathbf{W})^\top\bm{\beta}[t]+\mathbf{g}\right)\right)-\frac{1}{V_u^b}\frac{\mathbf{u}^B[t+1]}{t+1} \right.\notag\\
&\left.- \phi\left(\frac{1}{V_{u}^b}\left(\frac{1}{V_u}(\mathbf{M}\mathbf{W})^\top\bm{\beta}[t]+\mathbf{g}\right)\right) \right\rVert\notag\\
&+ \left\lVert\, \phi\left(\frac{1}{V_u^b}\left(\frac{t-1}{t}\frac{1}{V_u}(\mathbf{M}\mathbf{W})^\top\bm{\beta}[t-1]+\mathbf{g}\right)\right)-\frac{1}{V_u^b}\frac{\mathbf{u}^B[t]}{t} \right.\notag\\
&\left.- \phi\left(\frac{1}{V_{u}^b}\left(\frac{1}{V_u}(\mathbf{M}\mathbf{W})^\top\bm{\beta}[t-1]+\mathbf{g}\right)\right) \right\rVert\notag\\
\leq\quad&\left\lVert\, \phi\left(\frac{1}{V_{u}^b}\left(\frac{1}{V_u}(\mathbf{M}\mathbf{W})^\top\bm{\beta}[t]+\mathbf{g}\right)\right) - \phi\left(\frac{1}{V_{u}^b}\left(\frac{1}{V_u}(\mathbf{M}\mathbf{W})^\top\bm{\beta}[t-1]+\mathbf{g}\right)\right) \right\rVert\notag\\
&+ \frac{1}{V_{u}^b}\left(\left\lVert \frac{1}{t+1}\frac{1}{V_u}(\mathbf{M}\mathbf{W})^\top\bm{\beta}[t] \right\rVert + \left\lVert \frac{\mathbf{u}^B[t+1]}{t+1} \right\rVert \right.\notag\\
&\left.+  \left\lVert \frac{1}{t}\frac{1}{V_u}(\mathbf{M}\mathbf{W})^\top\bm{\beta}[t-1] \right\rVert + \left\lVert \frac{\mathbf{u}^B[t]}{t} \right\rVert \right).
\end{align}

As $\lVert (\mathbf{M}\mathbf{W})^\top \lVert_2 \leq \gamma V_{u}V_u^b, \gamma<1$, and $\lvert\mathbf{u}^B_i[t]\rvert$ is bounded, we have $\left\lVert \bm{\beta}[t+1] \right\rVert\leq \gamma \left\lVert \bm{\beta}[t] \right\rVert+\frac{1}{V_{u}}\left(\left\lVert \mathbf{g}\right\rVert+\left\lVert\frac{\mathbf{u}^B[t+1]}{t+1}\right\rVert\right)\leq \gamma \left\lVert \bm{\beta}[t] \right\rVert+c$, where $c$ is a constant. Therefore $\left\lVert \bm{\beta}[t] \right\rVert$ is bounded. Then $\forall \epsilon>0, \exists T_1$ such that when $t>T_1$, we have: \\
\begin{align}
    &\frac{1}{V_{u}^b}\left(\left\lVert \frac{1}{t+1}\frac{1}{V_u}(\mathbf{M}\mathbf{W})^\top\bm{\beta}[t] \right\rVert + \left\lVert \frac{\mathbf{u}^B[t+1]}{t+1} \right\rVert \right.\notag\\
    &\left.+  \left\lVert \frac{1}{t}\frac{1}{V_u}(\mathbf{M}\mathbf{W})^\top\bm{\beta}[t-1] \right\rVert + \left\lVert \frac{\mathbf{u}^B[t]}{t} \right\rVert \right) \leq \frac{\epsilon(1-\gamma)}{2}.
\end{align}

And since $\lVert (\mathbf{M}\mathbf{W})^\top \lVert_2 \leq \gamma V_{u}V_u^b$, we have: \\
\begin{align}
    &\left\lVert\, \phi\left(\frac{1}{V_{u}^b}\left(\frac{1}{V_u}(\mathbf{M}\mathbf{W})^\top\bm{\beta}[t]+\mathbf{g}\right)\right) - \phi\left(\frac{1}{V_{u}^b}\left(\frac{1}{V_u}(\mathbf{M}\mathbf{W})^\top\bm{\beta}[t-1]+\mathbf{g}\right)\right) \right\rVert \notag\\
    &\leq \gamma\left\lVert \bm{\beta}[t]-\bm{\beta}[t-1] \right\rVert.
\end{align}

Therefore, when $t>T_1$, it holds that:
\begin{equation}
    \left\lVert \bm{\beta}[t+1]-\bm{\beta}[t]\right\rVert \leq \gamma\left\lVert \bm{\beta}[t]-\bm{\beta}[t-1] \right\rVert + \frac{\epsilon(1-\gamma)}{2}.
\end{equation}

By iterating the above inequality, we have $ \lVert \bm{\beta}[t+1]-\bm{\beta}[t]\rVert \leq \gamma^{t-T_1}\lVert \bm{\beta}[T_1+1]-\bm{\beta}[T_1] \rVert + \frac{\epsilon(1-\gamma)}{2}\left(1+\gamma+\cdots+\gamma^{t-T_1-1}\right)<\gamma^{t-T_1}\lVert \bm{\beta}[T_1+1]-\bm{\beta}[T_1] \rVert + \frac{\epsilon}{2}.$ There exists $T_2$ such that when $t>T_1+T_2$, $\gamma^{t-T_1}\lVert \bm{\beta}[T_1+1]-\bm{\beta}[T_1] \rVert \leq \frac{\epsilon}{2}$, and therefore $\lVert \bm{\beta}[t+1]-\bm{\beta}[t]\rVert < \epsilon$.
According to Cauchy's convergence test, the sequence $\{\bm{\beta}[t]\}_{i=0}^{\infty}$ converges to $\bm{\beta}^*$. Considering the limit, it satisfies $\bm{\beta}^* = \phi\left(\frac{1}{V_{u}^b}\left(\frac{1}{V_u}(\mathbf{M}\mathbf{W})^\top\bm{\beta}^*+\mathbf{g}\right)\right)$.

When $V_{th}^b-u_{reset}^b=1$, and there exists $\lambda<1$ such that $\lVert(\mathbf{M}\mathbf{W})^\top\rVert_{\infty}\leq\lambda(V_{th}-u_{reset})$ and $\lVert\mathbf{g}\rVert_{\infty}\leq 1-\lambda$, the equation turns into $\bm{\beta}^* = \phi\left(\frac{1}{V_u}(\mathbf{M}\mathbf{W})^\top\bm{\beta}^*+\mathbf{g}\right)$. We have:
\begin{align}
    \left\lVert \bm{\beta}^* \right\rVert_{\infty} &= \left\lVert  \phi\left(\frac{1}{V_u}(\mathbf{M}\mathbf{W})^\top\bm{\beta}^*+\mathbf{g}\right) \right\rVert_{\infty} \notag\\
    &\leq \left\lVert  \frac{1}{V_u}(\mathbf{M}\mathbf{W})^\top\bm{\beta}^* \right\rVert_{\infty} + \left\lVert  \mathbf{g} \right\rVert_{\infty} \notag\\
    &\leq \lambda \left\lVert \bm{\beta}^* \right\rVert_{\infty} + \left\lVert  \mathbf{g} \right\rVert_{\infty}.
\end{align}

Therefore, $\left\lVert \bm{\beta}^* \right\rVert_{\infty}\leq \frac{\left\lVert  \mathbf{g} \right\rVert_{\infty}}{1-\lambda}\leq 1$, and we have $\left\lVert  \frac{1}{V_u}(\mathbf{M}\mathbf{W})^\top\bm{\beta}^*+\mathbf{g} \right\rVert_{\infty}\leq \\
\left\lVert  \frac{1}{V_u}(\mathbf{M}\mathbf{W})^\top\bm{\beta}^* \right\rVert_{\infty} + \left\lVert  \mathbf{g} \right\rVert_{\infty} \leq \lambda+(1-\lambda)=1$. It means $\bm{\beta}^* = \phi\left(\frac{1}{V_u}(\mathbf{M}\mathbf{W})^\top\bm{\beta}^*+\mathbf{g}\right) \\
= \frac{1}{V_u}(\mathbf{M}\mathbf{W})^\top\bm{\beta}^*+\mathbf{g}$.

Taking $f_{\theta}(\bm{\alpha}^*)=\sigma\left( \frac{1}{V_{th}-u_{reset}}(\mathbf{W}\bm{\alpha}^*+\mathbf{F}\mathbf{\mathbf{x}^*}+\mathbf{b}) \right)$ (i.e. the fixed-point equation at the equilibrium state as in Section~\ref{equilibrium states}) explicitly into Eq.~(\ref{eq:linear equation}), the linear equation turns into $\left( \frac{1}{V_u}(\mathbf{M}\mathbf{W})^\top-I \right)\bm{\beta}+\mathbf{g}=0$, where $V_u, \mathbf{M}, \mathbf{g}$ are previously defined. Therefore, $\bm{\beta}^*$ satisfies this equation. And since $\lVert (\mathbf{M}\mathbf{W})^\top \lVert_2 \leq \gamma V_{u}, \gamma<1$, the equation has the unique solution $\bm{\beta}^*$.

\end{proof}

\begin{remark}
As for the assumptions in the theorem, firstly, when $V_{th}^b-u_{reset}^b=1$ as we will take, the assumption for the convergence is weaker than that for the convergence in the forward stage (in Section~\ref{equilibrium states}), because $\lVert (\mathbf{M}\mathbf{W})^\top \rVert_2\leq \lVert \mathbf{W} \rVert_2$ as $\mathbf{M}$ is a diagonal mask matrix. We will restrict the spectral norm of $\mathbf{W}$ following \citet{xiao2021training} to encourage the convergence of the forward stage (in~\ref{appsec:training details}), then this backward stage would converge as well.

The assumptions for the consistency of the solution is a sufficient condition. In practice, the weight norm will be partially restricted by weight decay and our restriction on Frobenius norm (in~\ref{appsec:training details}), as well as the diagonal mask matrix $\mathbf{M}$ which would be sparse if the forward firing events are sparse, and we will rescale the loss so that the input $\mathbf{g}$ is in an appropriate range, as indicated in Section~\ref{subsec:error}. Even if these assumptions are not satisfied, we can view $\phi$ as a kind of empirical clipping techniques to stabilize the training, as indicated in Section~\ref{subsec:solve id}. The discussion is similar for the multi-layer condition (Theorem~\ref{thm2}) in the next section.
\end{remark}

\section{Proof of Theorem~\ref{thm2}}\label{appsec:proof2}

\begin{proof}

We first prove the convergence of $\bm{\beta}^l[t]$.
Let $V_u$ and $V_u^b$ denote $V_{th}-u_{reset}$ and $V_{th}^b-u_{reset}^b$ respectively. Let\\ $g_N^{t+1}(\bm{\beta}, \mathbf{g}, \mathbf{u}^B)=\phi\left(\frac{1}{V_{u}^b}\left(\frac{t}{t+1}\frac{1}{V_u}(\mathbf{M}^1\mathbf{W}^1)^\top\bm{\beta}+\mathbf{g}\right)\right)-\frac{1}{V_{u}^b}\frac{\mathbf{u}^B}{t+1}, \\g_{l}^{t}(\bm{\beta}, \mathbf{u}^B)=\phi\left(\frac{1}{V_{u}^b}\left(\frac{1}{V_u}(\mathbf{M}^{l+1}\mathbf{F}^{l+1})^\top\bm{\beta}\right)\right)-\frac{1}{V_{u}^b}\frac{\mathbf{u}^B}{t}, l=1,\cdots,N-1, \\g_N(\bm{\beta}, \mathbf{g})=\phi\left(\frac{1}{V_{u}^b}(\frac{1}{V_u}(\mathbf{M}^1\mathbf{W}^1)^\top\bm{\beta}+\mathbf{g})\right),\\ g_{l}(\bm{\beta}) = \phi\left(\frac{1}{V_{u}^b}\left(\frac{1}{V_u}(\mathbf{M}^{l+1}\mathbf{F}^{l+1})^\top\bm{\beta}\right)\right), l=1,\cdots,N-1.$
Then $\bm{\beta}^N[t+1] = g_N^{t+1}\left(g_1^{t}\left(\cdots g_{N-1}^{t}\left(\bm{\beta}^N[t], {\mathbf{u}^{N-1}}^B[t]\right) \cdots, {\mathbf{u}^1}^B[t]\right), \mathbf{g}, {\mathbf{u}^N}^B[t+1]\right)$.

We have:
\allowdisplaybreaks[4]
\begin{align}
& \left\lVert \bm{\beta}^N[t+1] - \bm{\beta}^N[t] \right\rVert \notag\\
=\quad & \left\lVert g_N^{t+1}\left(g_1^{t}\left(\cdots g_{N-1}^{t}\left(\bm{\beta}^N[t], {\mathbf{u}^{N-1}}^B[t]\right) \cdots, {\mathbf{u}^1}^B[t]\right), \mathbf{g}, {\mathbf{u}^N}^B[t+1]\right)\right. \notag\\
& \left.- g_N^{t}\left(g_1^{t-1}\left(\cdots g_{N-1}^{t-1}\left(\bm{\beta}^N[t-1], {\mathbf{u}^{N-1}}^B[t-1]\right) \cdots, {\mathbf{u}^1}^B[t-1]\right), \mathbf{g}, {\mathbf{u}^N}^B[t]\right) \right\rVert \notag\\ 
\leq\quad & \left\lVert g_N\left(g_1\left(\cdots g_{N-1}\left(\bm{\beta}^N[t]\right) \cdots\right), \mathbf{g}\right) - g_N\left(g_1\left(\cdots g_{N-1}\left(\bm{\beta}^N[t-1]\right) \cdots\right), \mathbf{g}\right) \right\rVert \notag\\
& + \left\lVert  g_N^{t+1}\left(g_1^{t}\left(\cdots g_{N-1}^{t}\left(\bm{\beta}^N[t], {\mathbf{u}^{N-1}}^B[t]\right) \cdots, {\mathbf{u}^1}^B[t]\right), \mathbf{g}, {\mathbf{u}^N}^B[t+1]\right)\right. \notag\\
&\left. - g_N\left(g_1\left(\cdots g_{N-1}\left(\bm{\beta}^N[t]\right) \cdots\right), \mathbf{g}\right) \right\rVert \notag\\
& + \left\lVert g_N^{t}\left(g_1^{t-1}\left(\cdots g_{N-1}^{t-1}\left(\bm{\beta}^N[t-1], {\mathbf{u}^{N-1}}^B[t-1]\right) \cdots, {\mathbf{u}^1}^B[t-1]\right), \mathbf{g}, {\mathbf{u}^N}^B[t]\right)\right. \notag\\
&\left. - g_N\left(g_1\left(\cdots g_{N-1}\left(\bm{\beta}^N[t-1]\right) \cdots\right), \mathbf{g}\right) \right\rVert \notag\\ 
\leq\quad & \left\lVert g_N\left(g_1\left(\cdots g_{N-1}\left(\bm{\beta}^N[t]\right) \cdots\right), \mathbf{g}\right) - g_N\left(g_1\left(\cdots g_{N-1}\left(\bm{\beta}^N[t-1]\right) \cdots\right), \mathbf{g}\right) \right\rVert \notag\\
& + \frac{1}{V_{u}^b}\left(\left\lVert \frac{1}{t+1}\frac{1}{V_u}(\mathbf{M}^1\mathbf{W}^1)^\top g_1^{t}\left(\cdots g_{N-1}^{t}\left(\bm{\beta}^N[t], {\mathbf{u}^{N-1}}^B[t]\right) \cdots, {\mathbf{u}^1}^B[t]\right) \right\rVert\right. \notag\\
& \left. +  \left\lVert \frac{1}{t}\frac{1}{V_u}(\mathbf{M}^1\mathbf{W}^1)^\top g_1^{t-1}\left(\cdots g_{N-1}^{t-1}\left(\bm{\beta}^N[t-1], {\mathbf{u}^{N-1}}^B[t-1]\right) \cdots, {\mathbf{u}^1}^B[t-1]\right) \right\rVert\right. \notag\\
& \left. + X + Y + \left\lVert \frac{{\mathbf{u}^N}^B[t+1]}{t+1} \right\rVert + \left\lVert \frac{{\mathbf{u}^N}^B[t]}{t} \right\rVert \right), 
\end{align}
where $X = \left\lVert \frac{1}{V_u}(\mathbf{M}^1\mathbf{W}^1)^\top\left(g_1^{t}\left(\cdots g_{N-1}^{t}\left(\bm{\beta}^N[t], {\mathbf{u}^{N-1}}^B[t]\right) \cdots, {\mathbf{u}^1}^B[t]\right) \right.\right.\\
\left.\left. - g_1\left(\cdots g_{N-1}\left(\bm{\beta}^N[t] \right) \cdots \right) \right) \right\rVert$, \\
and $Y=\left\lVert \frac{1}{V_u}(\mathbf{M}^1\mathbf{W}^1)^\top\left(g_1^{t-1}\left(\cdots g_{N-1}^{t-1}\left(\bm{\beta}^N[t-1], {\mathbf{u}^{N-1}}^B[t-1]\right) \cdots, {\mathbf{u}^1}^B[t-1]\right)\right.\right.\\
\left.\left. - g_1\left(\cdots g_{N-1}\left(\bm{\beta}^N[t-1] \right) \cdots \right) \right) \right\rVert$.

For the term $X$ and $Y$, they are bounded by:
\begin{equation}
\begin{aligned}
X\leq\quad & \frac{1}{V_{u}^b}\left(\left\lVert \frac{1}{V_u}(\mathbf{M}^1\mathbf{W}^1)^\top\frac{1}{V_u}(\mathbf{M}^N\mathbf{F}^N)^\top \left(g_{2}^t\left(\cdots g_{N-1}^t\left(\bm{\beta}^N[t], {\mathbf{u}^{N-1}}^B[t]  \right)\cdots, {\mathbf{u}^{2}}^B \right)\right.\right.\right.\\
&\left.\left.\left. - g_{2}\left(\cdots g_{N-1}\left(\bm{\beta}^N[t]\right)  \cdots\right) \right) \right\rVert + \left\lVert \frac{1}{V_u}(\mathbf{M}^1\mathbf{W}^1)^\top \frac{{\mathbf{u}^1}^B[t]}{t} \right\rVert \right)\\
\leq\quad & \cdots\cdots\\
\leq\quad & \frac{1}{V_{u}^b}\left\lVert \frac{1}{V_u}(\mathbf{M}^1\mathbf{W}^1)^\top \frac{{\mathbf{u}^1}^B[t]}{t} \right\rVert + \cdots \\
& + \frac{1}{{V_{u}^b}^{N-1}} \left\lVert \frac{1}{{V_u}^{N-1}}(\mathbf{M}^1\mathbf{W}^1)^\top(\mathbf{M}^N\mathbf{F}^N)^\top\cdots (\mathbf{M}^3\mathbf{F}^3)^\top\frac{{\mathbf{u}^{N-1}}^B[t]}{t} \right\rVert,
\end{aligned}
\end{equation}
and $Y$ has the same form as $X$ by substituting $t$ with $t-1$.

Since $\lVert (\mathbf{M}^1\mathbf{W}^1)^\top\rVert_2\lVert (\mathbf{M}^N\mathbf{F}^N)^\top\rVert_2\cdots\lVert (\mathbf{M}^2\mathbf{F}^2)^\top\rVert_2\leq \gamma V_{u}^N{V_{u}^b}^N$, we have:
\begin{equation}
\begin{aligned}
    & \left\lVert g_N\left(g_1\left(\cdots g_{N-1}\left(\bm{\beta}^N[t]\right) \cdots\right), \mathbf{g}\right) - g_N\left(g_1\left(\cdots g_{N-1}\left(\bm{\beta}^N[t-1]\right) \cdots\right), \mathbf{g}\right) \right\rVert\\
    \leq\quad & \left\lVert \frac{1}{V_{u}^b}\frac{1}{V_u}(\mathbf{M}^1\mathbf{W}^1)^\top\left(g_1\left(\cdots g_{N-1}\left(\bm{\beta}^N[t] \right) \cdots \right) - g_1\left(\cdots g_{N-1}\left(\bm{\beta}^N[t-1] \right) \cdots \right)\right) \right\rVert\\
    \leq\quad & \cdots\cdots\\
    \leq\quad & \left\lVert \frac{1}{{V_{u}^b}^N}\frac{1}{V_{u}^N} (\mathbf{M}^1\mathbf{W}^1)^\top(\mathbf{M}^N\mathbf{F}^N)^\top\cdots (\mathbf{M}^2\mathbf{F}^2)^\top \left(\bm{\beta}^N[t] - \bm{\beta}^N[t-1] \right) \right\rVert\\
    \leq\quad & \gamma \left\lVert \bm{\beta}^N[t] - \bm{\beta}^N[t-1] \right\rVert.
\end{aligned}
\end{equation}

And since ${\mathbf{u}_i^l}^B[t]$ is bounded, then $\forall\epsilon>0, \exists T_1$ such that when $t>T_1$, we have:
\begin{equation}
    \left\lVert \bm{\beta}^N[t+1] - \bm{\beta}^N[t] \right\rVert \leq \gamma \left\lVert \bm{\beta}^N[t] - \bm{\beta}^N[t-1] \right\rVert + \frac{\epsilon(1-\gamma)}{2}.
\end{equation}

Then $\lVert \bm{\beta}^N[t+1] - \bm{\beta}^N[t] \rVert < \gamma^{t-T_1}\lVert \bm{\beta}^N[T_1+1] - \bm{\beta}^N[T_1] \rVert+\frac{\epsilon}{2}$, and there exists $T_2$ such that when $t > T_1 + T_2$, $\lVert \bm{\beta}^N[t+1] - \bm{\beta}^N[t] \rVert < \epsilon$. According to Cauchy's convergence test, $\bm{\beta}^N[t]$ converges to ${\bm{\beta}^N}^*$, which satisfies ${\bm{\beta}^N}^* = g_N\left(g_1\circ\cdots\circ g_{N-1}({\bm{\beta}^N}^*), \mathbf{g}\right)$. Considering the limit, $\bm{\beta}^{l}[t]$ converges to ${\bm{\beta}^l}^*$, which satisfies ${\bm{\beta}^l}^*=g_{l}({\bm{\beta}^{l+1}}^*)$.

When $V_{th}^b-u_{reset}^b=1$, and there exists $\lambda<1$ such that $\lVert(\mathbf{M}^1\mathbf{W}^1)^\top\rVert_{\infty}\leq\lambda(V_{th}-u_{reset}), \lVert(\mathbf{M}^l\mathbf{F}^l)^\top\rVert_{\infty}\leq\lambda(V_{th}-u_{reset}), l=2,\cdots,N$ and $\lVert\mathbf{g}\rVert_{\infty}\leq 1-\lambda^N$, we have:
\begin{align}
    \left\lVert {\bm{\beta}^N}^* \right\rVert_{\infty} &= \left\lVert g_N\left(g_1\circ\cdots\circ g_{N-1}({\bm{\beta}^N}^*), \mathbf{g}\right) \right\rVert_{\infty} \notag\\
    &\leq \left\lVert \frac{1}{V_u}(\mathbf{M}^1\mathbf{W}^1)^\top g_1\circ\cdots\circ g_{N-1}({\bm{\beta}^N}^*) \right\rVert_{\infty} + \left\lVert \mathbf{g} \right\rVert_{\infty} \notag \\
    &\leq \lambda \left\lVert g_1\circ\cdots\circ g_{N-1}({\bm{\beta}^N}^*) \right\rVert_{\infty} + \left\lVert \mathbf{g} \right\rVert_{\infty} \notag\\
    &\leq \cdots\cdots \leq \lambda^N \left\lVert {\bm{\beta}^N}^* \right\rVert_{\infty} + \left\lVert \mathbf{g} \right\rVert_{\infty}
\end{align}

Therefore, $\left\lVert {\bm{\beta}^N}^* \right\rVert_{\infty} \leq \frac{\left\lVert \mathbf{g} \right\rVert_{\infty}}{1-\lambda^N}\leq 1$, and $\left\lVert \Tilde{g}_{N-1}({\bm{\beta}^N}^*) \right\rVert_{\infty}\leq \lambda \left\lVert {\bm{\beta}^N}^* \right\rVert_{\infty}\leq \lambda, \cdots\cdots, \left\lVert \Tilde{g}_1\circ\cdots\circ \Tilde{g}_{N-1}({\bm{\beta}^N}^*) \right\rVert_{\infty} \leq \lambda^{N-1}, \left\lVert \Tilde{g}_N\left(\Tilde{g}_1\circ\cdots\circ \Tilde{g}_{N-1}({\bm{\beta}^N}^*), \mathbf{g}\right) \right\rVert_{\infty} \leq \lambda^{N}+(1-\lambda^N)=1$, where $\Tilde{g}_N(\bm{\beta}, \mathbf{g})=\frac{1}{V_u}(\mathbf{M}^1\mathbf{W}^1)^\top\bm{\beta}+\mathbf{g}, \Tilde{g}_{l}(\bm{\beta}) = \frac{1}{V_u}(\mathbf{M}^{l+1}\mathbf{F}^{l+1})^\top\bm{\beta},\\
l=1,\cdots,N-1$, (i.e. $\Tilde{g}_l$ is $g_l$ without the function $\phi$). It means ${\bm{\beta}^N}^* = g_N\left(g_1\circ\cdots\circ g_{N-1}({\bm{\beta}^N}^*), \mathbf{g}\right) = \Tilde{g}_N\left(\Tilde{g}_1\circ\cdots\circ \Tilde{g}_{N-1}({\bm{\beta}^N}^*), \mathbf{g}\right)$ and ${\bm{\beta}^l}^*=g_{l}({\bm{\beta}^{l+1}}^*)=\Tilde{g}_{l}({\bm{\beta}^{l+1}}^*)$.

Taking ${\bm{\alpha}^1}^* = f_1\left(f_N\circ\cdots\circ f_2({\bm{\alpha}^1}^*), \mathbf{x}^*\right)$ and ${\bm{\alpha}^{l+1}}^*=f_{l+1}({\bm{\alpha}^l}^*)$ (i.e. the fixed-point equation at the equilibrium state as in Section~\ref{equilibrium states}) explicitly into Eq.~(\ref{eq:linear equation}), the linear equation turns into $\Tilde{g}_1\circ\cdots\circ \Tilde{g}_{N-1}({\bm{\beta}})-\bm{\beta}+\mathbf{g}=0$. Therefore, ${\bm{\beta}^N}^*$ satisfies this equation. And since $\lVert (\mathbf{M}^1\mathbf{W}^1)^\top\rVert_2\lVert (\mathbf{M}^N\mathbf{F}^N)^\top\rVert_2\cdots\lVert (\mathbf{M}^2\mathbf{F}^2)^\top\rVert_2\leq \gamma V_{u}^N, \gamma<1$, the equation has the unique solution ${\bm{\beta}^N}^*$. Further, because $\Tilde{g}_l(\bm{\beta})=\left(\frac{\partial h_{l+1}({\bm{\alpha}^N}^*)}{\partial h_l({\bm{\alpha}^N}^*)}\right)^{\top}\bm{\beta}$, where $h_l({\bm{\alpha}^N}^*)=f_l\circ\cdots\circ f_2\left(f_1({\bm{\alpha}^N}^*, \mathbf{x}^*)\right), l=N, \cdots, 1$, we have ${\bm{\beta}^l}^*=\left(\frac{\partial h_N({\bm{\alpha}^N}^*)}{\partial h_l({\bm{\alpha}^N}^*)}\right)^{\top}{\bm{\beta}^N}^*, l=N-1,\cdots,1$.

\end{proof}

\section{Training details}\label{appsec:training details}

\subsection{Dropout}

Dropout is a commonly used technique to prevent over-fitting, and we follow~\citet{bai2019deep,bai2020multiscale,xiao2021training} to leverage variational dropout, i.e. the dropout of each layer is the same at different time steps. Since applying dropout on the output of neurons is a linear operation with a mask and scaling factor, it can be integrated into the weight matrix without affecting the conclusions of convergence. The detailed computation with dropout is also illustrated in the pseudocode in Section~\ref{subsec:pipeline}.

\subsection{Restriction on weight norm}\label{appsubsec:restriction weight norm}

As indicated in the theorems, a sufficient condition for the convergence to equilibrium states in both forward and backward stages is the restriction on the weight spectral norm. \citet{xiao2021training} leverages re-parameterization to restrict the spectral norm, i.e. they re-parameterize $\mathbf{W}$ as $\mathbf{W} = \alpha \frac{\mathbf{W}}{\lVert \mathbf{W}\rVert_2}$, where $\lVert \mathbf{W}\rVert_2$ is computed as the implementation of Spectral Normalization and $\alpha$ is a learnable parameter to be clipped in the range of $[-c, c]$ ($c$ is a constant). However, the computation of spectral norm and re-parameterization may be unfriendly to neuromorphic computation. We adjust it for a more friendly calculation as follows.

First, the spectral norm is upper-bounded by the Frobenius norm: $\lVert \mathbf{W}\rVert_2\leq\lVert \mathbf{W}\rVert_F$. We can alternatively restrict the Frobenius norm which is easier to compute. Further, considering that connection weights may not be easy for readout compared with neuron outputs, we can approximate $\lVert \mathbf{W}\rVert_F$ by $\lVert \mathbf{W}\rVert_F=\sqrt{\text{tr}(\mathbf{W}\mathbf{W}^\top)}=\sqrt{\mathbb{E}_{\bm{\epsilon}\in \mathcal{N}(0, I_d)}\left[\lVert\bm{\epsilon}^\top\mathbf{W}\rVert_2^2\right]}$, according to the Hutchinson estimator~\citep{hutchinson1989stochastic}. It can be viewed as source neurons outputting noises and target neurons accumulating signals to estimate the Frobenius norm. We will estimate the norm based on the Monte-Carlo estimation (we will take 64 samples), which is similarly adopted by ~\citet{bai2021stabilizing} to estimate the norm of their Jacobian matrix. Then based on the estimation, we will restrict $\mathbf{W}$ as $\mathbf{W}=\alpha\frac{\mathbf{W}}{\lVert \mathbf{W}\rVert_F}$ where $\alpha=\min(c, \lVert \mathbf{W}\rVert_F)$, $c$ is a constant for norm range. This estimation and calculation may correspond to large amounts of noises in our brains, and a feedback inhibition on connection weights based on neuron outputs.

Following \citet{xiao2021training}, we only restrict the norm of feedback connection weight $\mathbf{W}^1$ for the multi-layer structure, which works well in practice.

\subsection{Other details}\label{appsec: other details}

\paragraph{Details on MNIST, CIFAR-10, and CIFAR-100}
For SNN models with feedback structure, we set $V_{th}=1, u_{reset}=-1$ in the forward stage to form an equivalent equilibrium state as \citet{xiao2021training}. The constant for restriction in~\ref{appsubsec:restriction weight norm} is $c=2$.
Following \citet{xiao2021training}, we train models by SGD with momentum for 100 epochs. The momentum is 0.9, the batch size is 128, and the initial learning rate is 0.05. For MNIST, the learning rate is decayed by 0.1 every 30 epochs, while for CIFAR-10 and CIFAR-100, it is decayed by 0.1 at the 50th and 75th epoch. We apply linear warmup for the learning rate in the first 400 iterations for CIFAR-10 and CIFAR-100. We apply the weight decay with $5\times 10^{-4}$ and variational dropout with the rate 0.2 for AlexNet-F and 0.25 for CIFARNet-F. The initialization of weights follows \citet{wu2018spatio}, i.e. we sample weights from the standard uniform distribution and normalize them on each output dimension. The scale for the loss function (as in Section~\ref{subsec:error}) is $100$ for MNIST, $400$ for CIFAR-10, and $500$ for CIFAR-100. 

For SNN models with degraded feedforward structure, our hyperparameters mostly follow \citet{thiele2019spikegrad}, i.e. we set $V_{th}=0.5, u_{reset}=-0.5$, train models by SGD with momentum 0.9 for 60 epochs, set batch size as 128, and the initial learning rate as 0.1 which is decayed by 0.1 every 20 epochs, and apply the variational dropout only on the first fully-connected layer with rate 0.5.

The notations for our structures mean: `64C5' represents a convolution operation with 64 output channels and kernel size 5, `s' after `64C5' means convolution with stride 2 (which downscales $2\times$) while `u' after that means a transposed convolution to upscale $2\times$, `P2' means average pooling with size 2, and `F' means feedback layers. The network structures for CIFAR-10 and CIFAR-100 are:

AlexNet~\citep{wu2019direct}: 96C3-256C3-P2-384C3-P2-384C3-256C3-1024-1024,

AlexNet-F~\citep{xiao2021training}: 96C3s-256C3-384C3s-384C3-256C3 (F96C3u),

CIFARNet~\citep{wu2019direct}: 128C3-256C3-P2-512C3-P2-1024C3-512C3-1024-512,

CIFARNet-F~\citep{xiao2021training}: 128C3s-256C3-512C3s-1024C3-512C3 (F128C3u).

\paragraph{Details on CIFAR10-DVS}
The CIFAR10-DVS dataset is the neuromorphic version of the CIFAR-10 dataset converted by a Dynamic Vision Sensor (DVS), which is composed of 10,000 samples, one-sixth of the original CIFAR-10. It consists of spike trains with two channels corresponding to ON- and OFF-event spikes. The pixel dimension is expanded to $128\times128$. Following the common practice, we split the dataset into 9000 training samples and 1000 testing samples. As for the data pre-processing, we reduce the time resolution by accumulating the spike events~\citep{Fang_2021_ICCV} into 30 time steps, and we reduce the spatial resolution into $48\times48$ by interpolation. We apply the same random crop augmentation as CIFAR-10 to the input data. We leverage the network structure: 512C9s (F512C5). We train the model by SGD with momentum for 70 epochs. The momentum is 0.9, the batch size is 128, the weight-decay is $5\times10^{-4}$, and the initial learning rate is 0.05 which is decayed by 0.1 at the 50th epoch. No dropout is applied. The initialization of weights follows the widely used Kaiming initialization. The constant for restriction in~\ref{appsubsec:restriction weight norm} is $c=10$ due to the large channel size, and the scale for the loss function as well as the firing thresholds and reset potentials are the same as the CIFAR-10 experiment.

We simulate the computation on commonly used computational units. The code implementation is based on the PyTorch framework~\citep{paszke2019pytorch}, and experiments are carried out on one NVIDIA GeForce RTX 3090 GPU.

\bibliography{SPIDE}

\begin{thebibliography}{78}
\expandafter\ifx\csname natexlab\endcsname\relax\def\natexlab#1{#1}\fi
\providecommand{\url}[1]{\texttt{#1}}
\providecommand{\href}[2]{#2}
\providecommand{\path}[1]{#1}
\providecommand{\DOIprefix}{doi:}
\providecommand{\ArXivprefix}{arXiv:}
\providecommand{\URLprefix}{URL: }
\providecommand{\Pubmedprefix}{pmid:}
\providecommand{\doi}[1]{\href{http://dx.doi.org/#1}{\path{#1}}}
\providecommand{\Pubmed}[1]{\href{pmid:#1}{\path{#1}}}
\providecommand{\bibinfo}[2]{#2}
\ifx\xfnm\relax \def\xfnm[#1]{\unskip,\space#1}\fi
\bibitem[{Akopyan et~al.(2015)Akopyan, Sawada, Cassidy, Alvarez-Icaza, Arthur,
  Merolla, Imam, Nakamura, Datta, Nam et~al.}]{akopyan2015truenorth}
\bibinfo{author}{Akopyan, F.}, \bibinfo{author}{Sawada, J.},
  \bibinfo{author}{Cassidy, A.}, \bibinfo{author}{Alvarez-Icaza, R.},
  \bibinfo{author}{Arthur, J.}, \bibinfo{author}{Merolla, P.},
  \bibinfo{author}{Imam, N.}, \bibinfo{author}{Nakamura, Y.},
  \bibinfo{author}{Datta, P.}, \bibinfo{author}{Nam, G.-J.} et~al.
  (\bibinfo{year}{2015}).
\newblock \bibinfo{title}{{TrueNorth: Design and tool flow of a 65 mw 1 million
  neuron programmable neurosynaptic chip}}.
\newblock {\it \bibinfo{journal}{IEEE Transactions on Computer-Aided Design of
  Integrated Circuits and Systems}\/},  {\it \bibinfo{volume}{34}\/},
  \bibinfo{pages}{1537--1557}.
\bibitem[{Almeida(1987)}]{almeida1987learning}
\bibinfo{author}{Almeida, L.} (\bibinfo{year}{1987}).
\newblock \bibinfo{title}{A learning rule for asynchronous perceptrons with
  feedback in a combinatorial environment.}
\newblock In {\it \bibinfo{booktitle}{International Conference on Neural
  Networks}\/}.
\bibitem[{Bai et~al.(2019)Bai, Kolter \& Koltun}]{bai2019deep}
\bibinfo{author}{Bai, S.}, \bibinfo{author}{Kolter, J.~Z.}, \&
  \bibinfo{author}{Koltun, V.} (\bibinfo{year}{2019}).
\newblock \bibinfo{title}{Deep equilibrium models}.
\newblock In {\it \bibinfo{booktitle}{Advances in Neural Information Processing
  Systems}\/}.
\bibitem[{Bai et~al.(2020)Bai, Koltun \& Kolter}]{bai2020multiscale}
\bibinfo{author}{Bai, S.}, \bibinfo{author}{Koltun, V.}, \&
  \bibinfo{author}{Kolter, J.~Z.} (\bibinfo{year}{2020}).
\newblock \bibinfo{title}{Multiscale deep equilibrium models}.
\newblock In {\it \bibinfo{booktitle}{Advances in Neural Information Processing
  Systems}\/}.
\bibitem[{Bai et~al.(2021)Bai, Koltun \& Kolter}]{bai2021stabilizing}
\bibinfo{author}{Bai, S.}, \bibinfo{author}{Koltun, V.}, \&
  \bibinfo{author}{Kolter, Z.} (\bibinfo{year}{2021}).
\newblock \bibinfo{title}{Stabilizing equilibrium models by jacobian
  regularization}.
\newblock In {\it \bibinfo{booktitle}{International Conference on Machine
  Learning}\/}.
\bibitem[{Bellec et~al.(2018)Bellec, Salaj, Subramoney, Legenstein \&
  Maass}]{bellec2018long}
\bibinfo{author}{Bellec, G.}, \bibinfo{author}{Salaj, D.},
  \bibinfo{author}{Subramoney, A.}, \bibinfo{author}{Legenstein, R.}, \&
  \bibinfo{author}{Maass, W.} (\bibinfo{year}{2018}).
\newblock \bibinfo{title}{Long short-term memory and learning-to-learn in
  networks of spiking neurons}.
\newblock In {\it \bibinfo{booktitle}{Advances in Neural Information Processing
  Systems}\/}.
\bibitem[{Bohte et~al.(2002)Bohte, Kok \& La~Poutre}]{bohte2002error}
\bibinfo{author}{Bohte, S.~M.}, \bibinfo{author}{Kok, J.~N.}, \&
  \bibinfo{author}{La~Poutre, H.} (\bibinfo{year}{2002}).
\newblock \bibinfo{title}{Error-backpropagation in temporally encoded networks
  of spiking neurons}.
\newblock {\it \bibinfo{journal}{Neurocomputing}\/},  {\it
  \bibinfo{volume}{48}\/}, \bibinfo{pages}{17--37}.
\bibitem[{Bottou(2010)}]{bottou2010large}
\bibinfo{author}{Bottou, L.} (\bibinfo{year}{2010}).
\newblock \bibinfo{title}{Large-scale machine learning with stochastic gradient
  descent}.
\newblock In {\it \bibinfo{booktitle}{Proceedings of COMPSTAT'2010}\/}.
\bibitem[{Brock et~al.(2021{\natexlab{a}})Brock, De \&
  Smith}]{brock2021characterizing}
\bibinfo{author}{Brock, A.}, \bibinfo{author}{De, S.}, \&
  \bibinfo{author}{Smith, S.~L.} (\bibinfo{year}{2021}{\natexlab{a}}).
\newblock \bibinfo{title}{Characterizing signal propagation to close the
  performance gap in unnormalized resnets}.
\newblock In {\it \bibinfo{booktitle}{International Conference on Learning
  Representations}\/}.
\bibitem[{Brock et~al.(2021{\natexlab{b}})Brock, De, Smith \&
  Simonyan}]{brock2021high}
\bibinfo{author}{Brock, A.}, \bibinfo{author}{De, S.}, \bibinfo{author}{Smith,
  S.~L.}, \& \bibinfo{author}{Simonyan, K.}
  (\bibinfo{year}{2021}{\natexlab{b}}).
\newblock \bibinfo{title}{High-performance large-scale image recognition
  without normalization}.
\newblock In {\it \bibinfo{booktitle}{International Conference on Machine
  Learning}\/}.
\bibitem[{Bu et~al.(2022)Bu, Fang, Ding, Dai, Yu \& Huang}]{bu2021optimal}
\bibinfo{author}{Bu, T.}, \bibinfo{author}{Fang, W.}, \bibinfo{author}{Ding,
  J.}, \bibinfo{author}{Dai, P.}, \bibinfo{author}{Yu, Z.}, \&
  \bibinfo{author}{Huang, T.} (\bibinfo{year}{2022}).
\newblock \bibinfo{title}{Optimal ann-snn conversion for high-accuracy and
  ultra-low-latency spiking neural networks}.
\newblock In {\it \bibinfo{booktitle}{International Conference on Learning
  Representations}\/}.
\bibitem[{Chen et~al.(2020)Chen, Gai, Yao, Mahoney \&
  Gonzalez}]{NEURIPS2020_099fe6b0}
\bibinfo{author}{Chen, J.}, \bibinfo{author}{Gai, Y.}, \bibinfo{author}{Yao,
  Z.}, \bibinfo{author}{Mahoney, M.~W.}, \& \bibinfo{author}{Gonzalez, J.~E.}
  (\bibinfo{year}{2020}).
\newblock \bibinfo{title}{A statistical framework for low-bitwidth training of
  deep neural networks}.
\newblock In {\it \bibinfo{booktitle}{Advances in Neural Information Processing
  Systems}\/}.
\bibitem[{Davies et~al.(2018)Davies, Srinivasa, Lin, Chinya, Cao, Choday,
  Dimou, Joshi, Imam, Jain et~al.}]{davies2018loihi}
\bibinfo{author}{Davies, M.}, \bibinfo{author}{Srinivasa, N.},
  \bibinfo{author}{Lin, T.-H.}, \bibinfo{author}{Chinya, G.},
  \bibinfo{author}{Cao, Y.}, \bibinfo{author}{Choday, S.~H.},
  \bibinfo{author}{Dimou, G.}, \bibinfo{author}{Joshi, P.},
  \bibinfo{author}{Imam, N.}, \bibinfo{author}{Jain, S.} et~al.
  (\bibinfo{year}{2018}).
\newblock \bibinfo{title}{Loihi: A neuromorphic manycore processor with on-chip
  learning}.
\newblock {\it \bibinfo{journal}{IEEE Micro}\/},  {\it \bibinfo{volume}{38}\/},
  \bibinfo{pages}{82--99}.
\bibitem[{Deng \& Gu(2021)}]{deng2021optimal}
\bibinfo{author}{Deng, S.}, \& \bibinfo{author}{Gu, S.} (\bibinfo{year}{2021}).
\newblock \bibinfo{title}{Optimal conversion of conventional artificial neural
  networks to spiking neural networks}.
\newblock In {\it \bibinfo{booktitle}{International Conference on Learning
  Representations}\/}.
\bibitem[{Deng et~al.(2022)Deng, Li, Zhang \& Gu}]{deng2021temporal}
\bibinfo{author}{Deng, S.}, \bibinfo{author}{Li, Y.}, \bibinfo{author}{Zhang,
  S.}, \& \bibinfo{author}{Gu, S.} (\bibinfo{year}{2022}).
\newblock \bibinfo{title}{Temporal efficient training of spiking neural network
  via gradient re-weighting}.
\newblock In {\it \bibinfo{booktitle}{International Conference on Learning
  Representations}\/}.
\bibitem[{Detorakis et~al.(2019)Detorakis, Bartley \&
  Neftci}]{detorakis2019contrastive}
\bibinfo{author}{Detorakis, G.}, \bibinfo{author}{Bartley, T.}, \&
  \bibinfo{author}{Neftci, E.} (\bibinfo{year}{2019}).
\newblock \bibinfo{title}{Contrastive hebbian learning with random feedback
  weights}.
\newblock {\it \bibinfo{journal}{Neural Networks}\/},  {\it
  \bibinfo{volume}{114}\/}, \bibinfo{pages}{1--14}.
\bibitem[{Diehl \& Cook(2015)}]{diehl2015unsupervised}
\bibinfo{author}{Diehl, P.~U.}, \& \bibinfo{author}{Cook, M.}
  (\bibinfo{year}{2015}).
\newblock \bibinfo{title}{Unsupervised learning of digit recognition using
  spike-timing-dependent plasticity}.
\newblock {\it \bibinfo{journal}{Frontiers in Computational Neuroscience}\/},
  {\it \bibinfo{volume}{9}\/}, \bibinfo{pages}{99}.
\bibitem[{Fang et~al.(2021{\natexlab{a}})Fang, Yu, Chen, Huang, Masquelier \&
  Tian}]{fang2021deep}
\bibinfo{author}{Fang, W.}, \bibinfo{author}{Yu, Z.}, \bibinfo{author}{Chen,
  Y.}, \bibinfo{author}{Huang, T.}, \bibinfo{author}{Masquelier, T.}, \&
  \bibinfo{author}{Tian, Y.} (\bibinfo{year}{2021}{\natexlab{a}}).
\newblock \bibinfo{title}{Deep residual learning in spiking neural networks}.
\newblock In {\it \bibinfo{booktitle}{Advances in Neural Information Processing
  Systems}\/}.
\bibitem[{Fang et~al.(2021{\natexlab{b}})Fang, Yu, Chen, Masquelier, Huang \&
  Tian}]{Fang_2021_ICCV}
\bibinfo{author}{Fang, W.}, \bibinfo{author}{Yu, Z.}, \bibinfo{author}{Chen,
  Y.}, \bibinfo{author}{Masquelier, T.}, \bibinfo{author}{Huang, T.}, \&
  \bibinfo{author}{Tian, Y.} (\bibinfo{year}{2021}{\natexlab{b}}).
\newblock \bibinfo{title}{Incorporating learnable membrane time constant to
  enhance learning of spiking neural networks}.
\newblock In {\it \bibinfo{booktitle}{Proceedings of the IEEE/CVF International
  Conference on Computer Vision (ICCV)}\/}.
\bibitem[{Guerguiev et~al.(2017)Guerguiev, Lillicrap \&
  Richards}]{guerguiev2017towards}
\bibinfo{author}{Guerguiev, J.}, \bibinfo{author}{Lillicrap, T.~P.}, \&
  \bibinfo{author}{Richards, B.~A.} (\bibinfo{year}{2017}).
\newblock \bibinfo{title}{Towards deep learning with segregated dendrites}.
\newblock {\it \bibinfo{journal}{Elife}\/},  {\it \bibinfo{volume}{6}\/},
  \bibinfo{pages}{e22901}.
\bibitem[{Hopfield(1982)}]{hopfield1982neural}
\bibinfo{author}{Hopfield, J.~J.} (\bibinfo{year}{1982}).
\newblock \bibinfo{title}{Neural networks and physical systems with emergent
  collective computational abilities}.
\newblock {\it \bibinfo{journal}{Proceedings of the National Academy of
  Sciences}\/},  {\it \bibinfo{volume}{79}\/}, \bibinfo{pages}{2554--2558}.
\bibitem[{Hopfield(1984)}]{hopfield1984neurons}
\bibinfo{author}{Hopfield, J.~J.} (\bibinfo{year}{1984}).
\newblock \bibinfo{title}{Neurons with graded response have collective
  computational properties like those of two-state neurons}.
\newblock {\it \bibinfo{journal}{Proceedings of the National Academy of
  Sciences}\/},  {\it \bibinfo{volume}{81}\/}, \bibinfo{pages}{3088--3092}.
\bibitem[{Hunsberger \& Eliasmith(2015)}]{hunsberger2015spiking}
\bibinfo{author}{Hunsberger, E.}, \& \bibinfo{author}{Eliasmith, C.}
  (\bibinfo{year}{2015}).
\newblock \bibinfo{title}{{Spiking deep networks with LIF neurons}}.
\newblock {\it \bibinfo{journal}{arXiv preprint arXiv:1510.08829}\/}, .
\bibitem[{Hutchinson(1989)}]{hutchinson1989stochastic}
\bibinfo{author}{Hutchinson, M.~F.} (\bibinfo{year}{1989}).
\newblock \bibinfo{title}{A stochastic estimator of the trace of the influence
  matrix for laplacian smoothing splines}.
\newblock {\it \bibinfo{journal}{Communications in Statistics-Simulation and
  Computation}\/},  {\it \bibinfo{volume}{18}\/}, \bibinfo{pages}{1059--1076}.
\bibitem[{Jin et~al.(2018)Jin, Zhang \& Li}]{jin2018hybrid}
\bibinfo{author}{Jin, Y.}, \bibinfo{author}{Zhang, W.}, \& \bibinfo{author}{Li,
  P.} (\bibinfo{year}{2018}).
\newblock \bibinfo{title}{Hybrid macro/micro level backpropagation for training
  deep spiking neural networks}.
\newblock In {\it \bibinfo{booktitle}{Advances in Neural Information Processing
  Systems}\/}.
\bibitem[{Kheradpisheh et~al.(2018)Kheradpisheh, Ganjtabesh, Thorpe \&
  Masquelier}]{kheradpisheh2018stdp}
\bibinfo{author}{Kheradpisheh, S.~R.}, \bibinfo{author}{Ganjtabesh, M.},
  \bibinfo{author}{Thorpe, S.~J.}, \& \bibinfo{author}{Masquelier, T.}
  (\bibinfo{year}{2018}).
\newblock \bibinfo{title}{Stdp-based spiking deep convolutional neural networks
  for object recognition}.
\newblock {\it \bibinfo{journal}{Neural Networks}\/},  {\it
  \bibinfo{volume}{99}\/}, \bibinfo{pages}{56--67}.
\bibitem[{Kim et~al.(2020{\natexlab{a}})Kim, Kim \& Kim}]{kim2020unifying}
\bibinfo{author}{Kim, J.}, \bibinfo{author}{Kim, K.}, \& \bibinfo{author}{Kim,
  J.-J.} (\bibinfo{year}{2020}{\natexlab{a}}).
\newblock \bibinfo{title}{Unifying activation-and timing-based learning rules
  for spiking neural networks}.
\newblock In {\it \bibinfo{booktitle}{Advances in Neural Information Processing
  Systems}\/}.
\bibitem[{Kim et~al.(2020{\natexlab{b}})Kim, Park, Na \& Yoon}]{kim2020spiking}
\bibinfo{author}{Kim, S.}, \bibinfo{author}{Park, S.}, \bibinfo{author}{Na,
  B.}, \& \bibinfo{author}{Yoon, S.} (\bibinfo{year}{2020}{\natexlab{b}}).
\newblock \bibinfo{title}{Spiking-yolo: spiking neural network for
  energy-efficient object detection}.
\newblock In {\it \bibinfo{booktitle}{Proceedings of the AAAI Conference on
  Artificial Intelligence}\/}.
\bibitem[{Kim et~al.(2022)Kim, Li, Park, Venkatesha \& Panda}]{kim2022neural}
\bibinfo{author}{Kim, Y.}, \bibinfo{author}{Li, Y.}, \bibinfo{author}{Park,
  H.}, \bibinfo{author}{Venkatesha, Y.}, \& \bibinfo{author}{Panda, P.}
  (\bibinfo{year}{2022}).
\newblock \bibinfo{title}{Neural architecture search for spiking neural
  networks}.
\newblock {\it \bibinfo{journal}{arXiv preprint arXiv:2201.10355}\/}, .
\bibitem[{Krizhevsky \& Hinton(2009)}]{krizhevsky2009learning}
\bibinfo{author}{Krizhevsky, A.}, \& \bibinfo{author}{Hinton, G.}
  (\bibinfo{year}{2009}).
\newblock {\it \bibinfo{title}{Learning multiple layers of features from tiny
  images}\/}.
\newblock \bibinfo{type}{Technical Report} University of Toronto.
\bibitem[{Kubilius et~al.(2019)Kubilius, Schrimpf, Kar, Rajalingham, Hong,
  Majaj, Issa, Bashivan, Prescott-Roy, Schmidt et~al.}]{kubilius2019brain}
\bibinfo{author}{Kubilius, J.}, \bibinfo{author}{Schrimpf, M.},
  \bibinfo{author}{Kar, K.}, \bibinfo{author}{Rajalingham, R.},
  \bibinfo{author}{Hong, H.}, \bibinfo{author}{Majaj, N.},
  \bibinfo{author}{Issa, E.}, \bibinfo{author}{Bashivan, P.},
  \bibinfo{author}{Prescott-Roy, J.}, \bibinfo{author}{Schmidt, K.} et~al.
  (\bibinfo{year}{2019}).
\newblock \bibinfo{title}{Brain-like object recognition with high-performing
  shallow recurrent anns}.
\newblock In {\it \bibinfo{booktitle}{Advances in Neural Information Processing
  Systems}\/}.
\bibitem[{LeCun et~al.(1998)LeCun, Bottou, Bengio \&
  Haffner}]{lecun1998gradient}
\bibinfo{author}{LeCun, Y.}, \bibinfo{author}{Bottou, L.},
  \bibinfo{author}{Bengio, Y.}, \& \bibinfo{author}{Haffner, P.}
  (\bibinfo{year}{1998}).
\newblock \bibinfo{title}{Gradient-based learning applied to document
  recognition}.
\newblock {\it \bibinfo{journal}{Proceedings of the IEEE}\/},  {\it
  \bibinfo{volume}{86}\/}, \bibinfo{pages}{2278--2324}.
\bibitem[{Lee et~al.(2016)Lee, Delbruck \& Pfeiffer}]{lee2016training}
\bibinfo{author}{Lee, J.~H.}, \bibinfo{author}{Delbruck, T.}, \&
  \bibinfo{author}{Pfeiffer, M.} (\bibinfo{year}{2016}).
\newblock \bibinfo{title}{Training deep spiking neural networks using
  backpropagation}.
\newblock {\it \bibinfo{journal}{Frontiers in Neuroscience}\/},  {\it
  \bibinfo{volume}{10}\/}, \bibinfo{pages}{508}.
\bibitem[{Legenstein et~al.(2008)Legenstein, Pecevski \&
  Maass}]{legenstein2008learning}
\bibinfo{author}{Legenstein, R.}, \bibinfo{author}{Pecevski, D.}, \&
  \bibinfo{author}{Maass, W.} (\bibinfo{year}{2008}).
\newblock \bibinfo{title}{A learning theory for reward-modulated
  spike-timing-dependent plasticity with application to biofeedback}.
\newblock {\it \bibinfo{journal}{PLoS Comput Biol}\/},  {\it
  \bibinfo{volume}{4}\/}, \bibinfo{pages}{e1000180}.
\bibitem[{Li et~al.(2017)Li, Liu, Ji, Li \& Shi}]{li2017cifar10}
\bibinfo{author}{Li, H.}, \bibinfo{author}{Liu, H.}, \bibinfo{author}{Ji, X.},
  \bibinfo{author}{Li, G.}, \& \bibinfo{author}{Shi, L.}
  (\bibinfo{year}{2017}).
\newblock \bibinfo{title}{Cifar10-dvs: an event-stream dataset for object
  classification}.
\newblock {\it \bibinfo{journal}{Frontiers in Neuroscience}\/},  {\it
  \bibinfo{volume}{11}\/}, \bibinfo{pages}{309}.
\bibitem[{Li \& Pehlevan(2020)}]{li2020minimax}
\bibinfo{author}{Li, Q.}, \& \bibinfo{author}{Pehlevan, C.}
  (\bibinfo{year}{2020}).
\newblock \bibinfo{title}{Minimax dynamics of optimally balanced spiking
  networks of excitatory and inhibitory neurons}.
\newblock In {\it \bibinfo{booktitle}{Advances in Neural Information Processing
  Systems}\/}.
\bibitem[{Li et~al.(2021{\natexlab{a}})Li, Deng, Dong, Gong \& Gu}]{li2021free}
\bibinfo{author}{Li, Y.}, \bibinfo{author}{Deng, S.}, \bibinfo{author}{Dong,
  X.}, \bibinfo{author}{Gong, R.}, \& \bibinfo{author}{Gu, S.}
  (\bibinfo{year}{2021}{\natexlab{a}}).
\newblock \bibinfo{title}{A free lunch from ann: Towards efficient, accurate
  spiking neural networks calibration}.
\newblock In {\it \bibinfo{booktitle}{International Conference on Machine
  Learning}\/}.
\bibitem[{Li et~al.(2021{\natexlab{b}})Li, Guo, Zhang, Deng, Hai \&
  Gu}]{li2021differentiable}
\bibinfo{author}{Li, Y.}, \bibinfo{author}{Guo, Y.}, \bibinfo{author}{Zhang,
  S.}, \bibinfo{author}{Deng, S.}, \bibinfo{author}{Hai, Y.}, \&
  \bibinfo{author}{Gu, S.} (\bibinfo{year}{2021}{\natexlab{b}}).
\newblock \bibinfo{title}{Differentiable spike: Rethinking gradient-descent for
  training spiking neural networks}.
\newblock In {\it \bibinfo{booktitle}{Advances in Neural Information Processing
  Systems}\/}.
\bibitem[{Maass(1997)}]{maass1997networks}
\bibinfo{author}{Maass, W.} (\bibinfo{year}{1997}).
\newblock \bibinfo{title}{Networks of spiking neurons: the third generation of
  neural network models}.
\newblock {\it \bibinfo{journal}{Neural Networks}\/},  {\it
  \bibinfo{volume}{10}\/}, \bibinfo{pages}{1659--1671}.
\bibitem[{Mancoo et~al.(2020)Mancoo, Keemink \&
  Machens}]{mancoo2020understanding}
\bibinfo{author}{Mancoo, A.}, \bibinfo{author}{Keemink, S.}, \&
  \bibinfo{author}{Machens, C.~K.} (\bibinfo{year}{2020}).
\newblock \bibinfo{title}{Understanding spiking networks through convex
  optimization}.
\newblock In {\it \bibinfo{booktitle}{Advances in Neural Information Processing
  Systems}\/}.
\bibitem[{Martin et~al.(2021)Martin, Ernoult, Laydevant, Li, Querlioz, Petrisor
  \& Grollier}]{martin2021eqspike}
\bibinfo{author}{Martin, E.}, \bibinfo{author}{Ernoult, M.},
  \bibinfo{author}{Laydevant, J.}, \bibinfo{author}{Li, S.},
  \bibinfo{author}{Querlioz, D.}, \bibinfo{author}{Petrisor, T.}, \&
  \bibinfo{author}{Grollier, J.} (\bibinfo{year}{2021}).
\newblock \bibinfo{title}{Eqspike: spike-driven equilibrium propagation for
  neuromorphic implementations}.
\newblock {\it \bibinfo{journal}{Iscience}\/},  {\it \bibinfo{volume}{24}\/},
  \bibinfo{pages}{102222}.
\bibitem[{Meng et~al.(2022{\natexlab{a}})Meng, Xiao, Yan, Wang, Lin \&
  Luo}]{meng2022training}
\bibinfo{author}{Meng, Q.}, \bibinfo{author}{Xiao, M.}, \bibinfo{author}{Yan,
  S.}, \bibinfo{author}{Wang, Y.}, \bibinfo{author}{Lin, Z.}, \&
  \bibinfo{author}{Luo, Z.-Q.} (\bibinfo{year}{2022}{\natexlab{a}}).
\newblock \bibinfo{title}{Training high-performance low-latency spiking neural
  networks by differentiation on spike representation}.
\newblock In {\it \bibinfo{booktitle}{Proceedings of the IEEE Conference on
  Computer Vision and Pattern Recognition}\/}.
\bibitem[{Meng et~al.(2022{\natexlab{b}})Meng, Yan, Xiao, Wang, Lin \&
  Luo}]{meng2022trainingnn}
\bibinfo{author}{Meng, Q.}, \bibinfo{author}{Yan, S.}, \bibinfo{author}{Xiao,
  M.}, \bibinfo{author}{Wang, Y.}, \bibinfo{author}{Lin, Z.}, \&
  \bibinfo{author}{Luo, Z.-Q.} (\bibinfo{year}{2022}{\natexlab{b}}).
\newblock \bibinfo{title}{Training much deeper spiking neural networks with a
  small number of time-steps}.
\newblock {\it \bibinfo{journal}{Neural Networks}\/},  {\it
  \bibinfo{volume}{153}\/}, \bibinfo{pages}{254--268}.
\bibitem[{Mesnard et~al.(2016)Mesnard, Gerstner \& Brea}]{mesnard2016towards}
\bibinfo{author}{Mesnard, T.}, \bibinfo{author}{Gerstner, W.}, \&
  \bibinfo{author}{Brea, J.} (\bibinfo{year}{2016}).
\newblock \bibinfo{title}{Towards deep learning with spiking neurons in energy
  based models with contrastive hebbian plasticity}.
\newblock {\it \bibinfo{journal}{arXiv preprint arXiv:1612.03214}\/}, .
\bibitem[{Neftci et~al.(2017)Neftci, Augustine, Paul \&
  Detorakis}]{neftci2017event}
\bibinfo{author}{Neftci, E.~O.}, \bibinfo{author}{Augustine, C.},
  \bibinfo{author}{Paul, S.}, \& \bibinfo{author}{Detorakis, G.}
  (\bibinfo{year}{2017}).
\newblock \bibinfo{title}{Event-driven random back-propagation: Enabling
  neuromorphic deep learning machines}.
\newblock {\it \bibinfo{journal}{Frontiers in neuroscience}\/},  {\it
  \bibinfo{volume}{11}\/}, \bibinfo{pages}{324}.
\bibitem[{Neftci et~al.(2019)Neftci, Mostafa \& Zenke}]{neftci2019surrogate}
\bibinfo{author}{Neftci, E.~O.}, \bibinfo{author}{Mostafa, H.}, \&
  \bibinfo{author}{Zenke, F.} (\bibinfo{year}{2019}).
\newblock \bibinfo{title}{Surrogate gradient learning in spiking neural
  networks: Bringing the power of gradient-based optimization to spiking neural
  networks}.
\newblock {\it \bibinfo{journal}{IEEE Signal Processing Magazine}\/},  {\it
  \bibinfo{volume}{36}\/}, \bibinfo{pages}{51--63}.
\bibitem[{N{\o}kland(2016)}]{nokland2016direct}
\bibinfo{author}{N{\o}kland, A.} (\bibinfo{year}{2016}).
\newblock \bibinfo{title}{Direct feedback alignment provides learning in deep
  neural networks}.
\newblock In {\it \bibinfo{booktitle}{Advances in Neural Information Processing
  Systems}\/}.
\bibitem[{O’Connor et~al.(2019)O’Connor, Gavves \& Welling}]{o2019training}
\bibinfo{author}{O’Connor, P.}, \bibinfo{author}{Gavves, E.}, \&
  \bibinfo{author}{Welling, M.} (\bibinfo{year}{2019}).
\newblock \bibinfo{title}{Training a spiking neural network with equilibrium
  propagation}.
\newblock In {\it \bibinfo{booktitle}{The 22nd International Conference on
  Artificial Intelligence and Statistics}\/}.
\bibitem[{O'Connor \& Welling(2016)}]{o2016deep}
\bibinfo{author}{O'Connor, P.}, \& \bibinfo{author}{Welling, M.}
  (\bibinfo{year}{2016}).
\newblock \bibinfo{title}{Deep spiking networks}.
\newblock {\it \bibinfo{journal}{arXiv preprint arXiv:1602.08323}\/}, .
\bibitem[{Paszke et~al.(2019)Paszke, Gross, Massa, Lerer, Bradbury, Chanan,
  Killeen, Lin, Gimelshein, Antiga et~al.}]{paszke2019pytorch}
\bibinfo{author}{Paszke, A.}, \bibinfo{author}{Gross, S.},
  \bibinfo{author}{Massa, F.}, \bibinfo{author}{Lerer, A.},
  \bibinfo{author}{Bradbury, J.}, \bibinfo{author}{Chanan, G.},
  \bibinfo{author}{Killeen, T.}, \bibinfo{author}{Lin, Z.},
  \bibinfo{author}{Gimelshein, N.}, \bibinfo{author}{Antiga, L.} et~al.
  (\bibinfo{year}{2019}).
\newblock \bibinfo{title}{Pytorch: An imperative style, high-performance deep
  learning library}.
\newblock In {\it \bibinfo{booktitle}{Advances in Neural Information Processing
  Systems}\/}.
\bibitem[{Pei et~al.(2019)Pei, Deng, Song, Zhao, Zhang, Wu, Wang, Zou, Wu, He
  et~al.}]{pei2019towards}
\bibinfo{author}{Pei, J.}, \bibinfo{author}{Deng, L.}, \bibinfo{author}{Song,
  S.}, \bibinfo{author}{Zhao, M.}, \bibinfo{author}{Zhang, Y.},
  \bibinfo{author}{Wu, S.}, \bibinfo{author}{Wang, G.}, \bibinfo{author}{Zou,
  Z.}, \bibinfo{author}{Wu, Z.}, \bibinfo{author}{He, W.} et~al.
  (\bibinfo{year}{2019}).
\newblock \bibinfo{title}{{Towards artificial general intelligence with hybrid
  Tianjic chip architecture}}.
\newblock {\it \bibinfo{journal}{Nature}\/},  {\it \bibinfo{volume}{572}\/},
  \bibinfo{pages}{106--111}.
\bibitem[{Pineda(1987)}]{pineda1987generalization}
\bibinfo{author}{Pineda, F.~J.} (\bibinfo{year}{1987}).
\newblock \bibinfo{title}{Generalization of back-propagation to recurrent
  neural networks}.
\newblock {\it \bibinfo{journal}{Physical Review Letters}\/},  {\it
  \bibinfo{volume}{59}\/}, \bibinfo{pages}{2229}.
\bibitem[{Rathi et~al.(2020)Rathi, Srinivasan, Panda \&
  Roy}]{rathi2019enabling}
\bibinfo{author}{Rathi, N.}, \bibinfo{author}{Srinivasan, G.},
  \bibinfo{author}{Panda, P.}, \& \bibinfo{author}{Roy, K.}
  (\bibinfo{year}{2020}).
\newblock \bibinfo{title}{Enabling deep spiking neural networks with hybrid
  conversion and spike timing dependent backpropagation}.
\newblock In {\it \bibinfo{booktitle}{International Conference on Learning
  Representations}\/}.
\bibitem[{Roy et~al.(2019)Roy, Jaiswal \& Panda}]{roy2019towards}
\bibinfo{author}{Roy, K.}, \bibinfo{author}{Jaiswal, A.}, \&
  \bibinfo{author}{Panda, P.} (\bibinfo{year}{2019}).
\newblock \bibinfo{title}{Towards spike-based machine intelligence with
  neuromorphic computing}.
\newblock {\it \bibinfo{journal}{Nature}\/},  {\it \bibinfo{volume}{575}\/},
  \bibinfo{pages}{607--617}.
\bibitem[{Rueckauer et~al.(2017)Rueckauer, Lungu, Hu, Pfeiffer \&
  Liu}]{rueckauer2017conversion}
\bibinfo{author}{Rueckauer, B.}, \bibinfo{author}{Lungu, I.-A.},
  \bibinfo{author}{Hu, Y.}, \bibinfo{author}{Pfeiffer, M.}, \&
  \bibinfo{author}{Liu, S.-C.} (\bibinfo{year}{2017}).
\newblock \bibinfo{title}{Conversion of continuous-valued deep networks to
  efficient event-driven networks for image classification}.
\newblock {\it \bibinfo{journal}{Frontiers in Neuroscience}\/},  {\it
  \bibinfo{volume}{11}\/}, \bibinfo{pages}{682}.
\bibitem[{Rumelhart et~al.(1986)Rumelhart, Hinton \&
  Williams}]{rumelhart1986learning}
\bibinfo{author}{Rumelhart, D.~E.}, \bibinfo{author}{Hinton, G.~E.}, \&
  \bibinfo{author}{Williams, R.~J.} (\bibinfo{year}{1986}).
\newblock \bibinfo{title}{Learning representations by back-propagating errors}.
\newblock {\it \bibinfo{journal}{Nature}\/},  {\it \bibinfo{volume}{323}\/},
  \bibinfo{pages}{533--536}.
\bibitem[{Samadi et~al.(2017)Samadi, Lillicrap \& Tweed}]{samadi2017deep}
\bibinfo{author}{Samadi, A.}, \bibinfo{author}{Lillicrap, T.~P.}, \&
  \bibinfo{author}{Tweed, D.~B.} (\bibinfo{year}{2017}).
\newblock \bibinfo{title}{Deep learning with dynamic spiking neurons and fixed
  feedback weights}.
\newblock {\it \bibinfo{journal}{Neural Computation}\/},  {\it
  \bibinfo{volume}{29}\/}, \bibinfo{pages}{578--602}.
\bibitem[{Scellier \& Bengio(2017)}]{scellier2017equilibrium}
\bibinfo{author}{Scellier, B.}, \& \bibinfo{author}{Bengio, Y.}
  (\bibinfo{year}{2017}).
\newblock \bibinfo{title}{Equilibrium propagation: Bridging the gap between
  energy-based models and backpropagation}.
\newblock {\it \bibinfo{journal}{Frontiers in Computational Neuroscience}\/},
  {\it \bibinfo{volume}{11}\/}, \bibinfo{pages}{24}.
\bibitem[{Sengupta et~al.(2019)Sengupta, Ye, Wang, Liu \&
  Roy}]{sengupta2019going}
\bibinfo{author}{Sengupta, A.}, \bibinfo{author}{Ye, Y.},
  \bibinfo{author}{Wang, R.}, \bibinfo{author}{Liu, C.}, \&
  \bibinfo{author}{Roy, K.} (\bibinfo{year}{2019}).
\newblock \bibinfo{title}{Going deeper in spiking neural networks: Vgg and
  residual architectures}.
\newblock {\it \bibinfo{journal}{Frontiers in Neuroscience}\/},  {\it
  \bibinfo{volume}{13}\/}, \bibinfo{pages}{95}.
\bibitem[{Shrestha \& Orchard(2018)}]{shrestha2018slayer}
\bibinfo{author}{Shrestha, S.~B.}, \& \bibinfo{author}{Orchard, G.}
  (\bibinfo{year}{2018}).
\newblock \bibinfo{title}{Slayer: spike layer error reassignment in time}.
\newblock In {\it \bibinfo{booktitle}{Advances in Neural Information Processing
  Systems}\/}.
\bibitem[{Sironi et~al.(2018)Sironi, Brambilla, Bourdis, Lagorce \&
  Benosman}]{sironi2018hats}
\bibinfo{author}{Sironi, A.}, \bibinfo{author}{Brambilla, M.},
  \bibinfo{author}{Bourdis, N.}, \bibinfo{author}{Lagorce, X.}, \&
  \bibinfo{author}{Benosman, R.} (\bibinfo{year}{2018}).
\newblock \bibinfo{title}{Hats: Histograms of averaged time surfaces for robust
  event-based object classification}.
\newblock In {\it \bibinfo{booktitle}{Proceedings of the IEEE Conference on
  Computer Vision and Pattern Recognition}\/}.
\bibitem[{St{\"o}ckl \& Maass(2021)}]{stockl2021optimized}
\bibinfo{author}{St{\"o}ckl, C.}, \& \bibinfo{author}{Maass, W.}
  (\bibinfo{year}{2021}).
\newblock \bibinfo{title}{Optimized spiking neurons can classify images with
  high accuracy through temporal coding with two spikes}.
\newblock {\it \bibinfo{journal}{Nature Machine Intelligence}\/},  {\it
  \bibinfo{volume}{3}\/}, \bibinfo{pages}{230--238}.
\bibitem[{Tavanaei et~al.(2019)Tavanaei, Ghodrati, Kheradpisheh, Masquelier \&
  Maida}]{tavanaei2019deep}
\bibinfo{author}{Tavanaei, A.}, \bibinfo{author}{Ghodrati, M.},
  \bibinfo{author}{Kheradpisheh, S.~R.}, \bibinfo{author}{Masquelier, T.}, \&
  \bibinfo{author}{Maida, A.} (\bibinfo{year}{2019}).
\newblock \bibinfo{title}{Deep learning in spiking neural networks}.
\newblock {\it \bibinfo{journal}{Neural Networks}\/},  {\it
  \bibinfo{volume}{111}\/}, \bibinfo{pages}{47--63}.
\bibitem[{Thiele et~al.(2020)Thiele, Bichler \& Dupret}]{thiele2019spikegrad}
\bibinfo{author}{Thiele, J.~C.}, \bibinfo{author}{Bichler, O.}, \&
  \bibinfo{author}{Dupret, A.} (\bibinfo{year}{2020}).
\newblock \bibinfo{title}{Spikegrad: An ann-equivalent computation model for
  implementing backpropagation with spikes}.
\newblock In {\it \bibinfo{booktitle}{International Conference on Learning
  Representations}\/}.
\bibitem[{Thiele et~al.(2019)Thiele, Bichler, Dupret, Solinas \&
  Indiveri}]{thiele2019spiking}
\bibinfo{author}{Thiele, J.~C.}, \bibinfo{author}{Bichler, O.},
  \bibinfo{author}{Dupret, A.}, \bibinfo{author}{Solinas, S.}, \&
  \bibinfo{author}{Indiveri, G.} (\bibinfo{year}{2019}).
\newblock \bibinfo{title}{A spiking network for inference of relations trained
  with neuromorphic backpropagation}.
\newblock In {\it \bibinfo{booktitle}{2019 International Joint Conference on
  Neural Networks (IJCNN)}\/}.
\bibitem[{Wu et~al.(2021{\natexlab{a}})Wu, Zhang, Weng, Zhang, Xiong, Zha, Sun
  \& Wu}]{wu2021training}
\bibinfo{author}{Wu, H.}, \bibinfo{author}{Zhang, Y.}, \bibinfo{author}{Weng,
  W.}, \bibinfo{author}{Zhang, Y.}, \bibinfo{author}{Xiong, Z.},
  \bibinfo{author}{Zha, Z.-J.}, \bibinfo{author}{Sun, X.}, \&
  \bibinfo{author}{Wu, F.} (\bibinfo{year}{2021}{\natexlab{a}}).
\newblock \bibinfo{title}{Training spiking neural networks with accumulated
  spiking flow}.
\newblock In {\it \bibinfo{booktitle}{Proceedings of the AAAI Conference on
  Artificial Intelligence}\/}.
\bibitem[{Wu et~al.(2021{\natexlab{b}})Wu, Chua, Zhang, Li, Li \&
  Tan}]{wu2021tandem}
\bibinfo{author}{Wu, J.}, \bibinfo{author}{Chua, Y.}, \bibinfo{author}{Zhang,
  M.}, \bibinfo{author}{Li, G.}, \bibinfo{author}{Li, H.}, \&
  \bibinfo{author}{Tan, K.~C.} (\bibinfo{year}{2021}{\natexlab{b}}).
\newblock \bibinfo{title}{A tandem learning rule for effective training and
  rapid inference of deep spiking neural networks}.
\newblock {\it \bibinfo{journal}{IEEE Transactions on Neural Networks and
  Learning Systems}\/},  (pp. \bibinfo{pages}{1--15}).
\bibitem[{Wu et~al.(2018)Wu, Deng, Li, Zhu \& Shi}]{wu2018spatio}
\bibinfo{author}{Wu, Y.}, \bibinfo{author}{Deng, L.}, \bibinfo{author}{Li, G.},
  \bibinfo{author}{Zhu, J.}, \& \bibinfo{author}{Shi, L.}
  (\bibinfo{year}{2018}).
\newblock \bibinfo{title}{Spatio-temporal backpropagation for training
  high-performance spiking neural networks}.
\newblock {\it \bibinfo{journal}{Frontiers in Neuroscience}\/},  {\it
  \bibinfo{volume}{12}\/}, \bibinfo{pages}{331}.
\bibitem[{Wu et~al.(2019)Wu, Deng, Li, Zhu, Xie \& Shi}]{wu2019direct}
\bibinfo{author}{Wu, Y.}, \bibinfo{author}{Deng, L.}, \bibinfo{author}{Li, G.},
  \bibinfo{author}{Zhu, J.}, \bibinfo{author}{Xie, Y.}, \&
  \bibinfo{author}{Shi, L.} (\bibinfo{year}{2019}).
\newblock \bibinfo{title}{Direct training for spiking neural networks: Faster,
  larger, better}.
\newblock In {\it \bibinfo{booktitle}{Proceedings of the AAAI Conference on
  Artificial Intelligence}\/}.
\bibitem[{Xiao et~al.(2022)Xiao, Meng, Zhang, He \& Lin}]{xiao2022online}
\bibinfo{author}{Xiao, M.}, \bibinfo{author}{Meng, Q.}, \bibinfo{author}{Zhang,
  Z.}, \bibinfo{author}{He, D.}, \& \bibinfo{author}{Lin, Z.}
  (\bibinfo{year}{2022}).
\newblock \bibinfo{title}{Online training through time for spiking neural
  networks}.
\newblock In {\it \bibinfo{booktitle}{Advances in Neural Information Processing
  Systems}\/}.
\bibitem[{Xiao et~al.(2021)Xiao, Meng, Zhang, Wang \& Lin}]{xiao2021training}
\bibinfo{author}{Xiao, M.}, \bibinfo{author}{Meng, Q.}, \bibinfo{author}{Zhang,
  Z.}, \bibinfo{author}{Wang, Y.}, \& \bibinfo{author}{Lin, Z.}
  (\bibinfo{year}{2021}).
\newblock \bibinfo{title}{Training feedback spiking neural networks by implicit
  differentiation on the equilibrium state}.
\newblock In {\it \bibinfo{booktitle}{Advances in Neural Information Processing
  Systems}\/}.
\bibitem[{Xie \& Seung(2003)}]{xie2003equivalence}
\bibinfo{author}{Xie, X.}, \& \bibinfo{author}{Seung, H.~S.}
  (\bibinfo{year}{2003}).
\newblock \bibinfo{title}{Equivalence of backpropagation and contrastive
  hebbian learning in a layered network}.
\newblock {\it \bibinfo{journal}{Neural Computation}\/},  {\it
  \bibinfo{volume}{15}\/}, \bibinfo{pages}{441--454}.
\bibitem[{Yan et~al.(2021)Yan, Zhou \& Wong}]{yan2021near}
\bibinfo{author}{Yan, Z.}, \bibinfo{author}{Zhou, J.}, \&
  \bibinfo{author}{Wong, W.-F.} (\bibinfo{year}{2021}).
\newblock \bibinfo{title}{Near lossless transfer learning for spiking neural
  networks}.
\newblock In {\it \bibinfo{booktitle}{Proceedings of the AAAI Conference on
  Artificial Intelligence}\/}.
\bibitem[{Zhang et~al.(2018{\natexlab{a}})Zhang, Zeng, Zhao \&
  Shi}]{zhang2018plasticity}
\bibinfo{author}{Zhang, T.}, \bibinfo{author}{Zeng, Y.}, \bibinfo{author}{Zhao,
  D.}, \& \bibinfo{author}{Shi, M.} (\bibinfo{year}{2018}{\natexlab{a}}).
\newblock \bibinfo{title}{A plasticity-centric approach to train the
  non-differential spiking neural networks}.
\newblock In {\it \bibinfo{booktitle}{Proceedings of the AAAI Conference on
  Artificial Intelligence}\/}.
\bibitem[{Zhang et~al.(2018{\natexlab{b}})Zhang, Zeng, Zhao \&
  Xu}]{zhang2018brain}
\bibinfo{author}{Zhang, T.}, \bibinfo{author}{Zeng, Y.}, \bibinfo{author}{Zhao,
  D.}, \& \bibinfo{author}{Xu, B.} (\bibinfo{year}{2018}{\natexlab{b}}).
\newblock \bibinfo{title}{Brain-inspired balanced tuning for spiking neural
  networks.}
\newblock In {\it \bibinfo{booktitle}{Proceedings of the International Joint
  Conference on Artificial Intelligence}\/}.
\bibitem[{Zhang \& Li(2019)}]{zhang2019spike}
\bibinfo{author}{Zhang, W.}, \& \bibinfo{author}{Li, P.}
  (\bibinfo{year}{2019}).
\newblock \bibinfo{title}{Spike-train level backpropagation for training deep
  recurrent spiking neural networks}.
\newblock In {\it \bibinfo{booktitle}{Advances in Neural Information Processing
  Systems}\/}.
\bibitem[{Zhang \& Li(2020)}]{zhang2020temporal}
\bibinfo{author}{Zhang, W.}, \& \bibinfo{author}{Li, P.}
  (\bibinfo{year}{2020}).
\newblock \bibinfo{title}{Temporal spike sequence learning via backpropagation
  for deep spiking neural networks}.
\newblock In {\it \bibinfo{booktitle}{Advances in Neural Information Processing
  Systems}\/}.
\bibitem[{Zheng et~al.(2021)Zheng, Wu, Deng, Hu \& Li}]{zheng2020going}
\bibinfo{author}{Zheng, H.}, \bibinfo{author}{Wu, Y.}, \bibinfo{author}{Deng,
  L.}, \bibinfo{author}{Hu, Y.}, \& \bibinfo{author}{Li, G.}
  (\bibinfo{year}{2021}).
\newblock \bibinfo{title}{Going deeper with directly-trained larger spiking
  neural networks}.
\newblock In {\it \bibinfo{booktitle}{Proceedings of the AAAI Conference on
  Artificial Intelligence}\/}.

\end{thebibliography}
\end{document}